\documentclass{article} 
\usepackage{iclr2024_conference,times}

\usepackage{amsmath}
\usepackage[utf8]{inputenc} 
\usepackage[T1]{fontenc}    
\usepackage{hyperref}       
\usepackage{url}            
\usepackage{booktabs}       
\usepackage{amsfonts}       
\usepackage{nicefrac}       
\usepackage{multirow}
\usepackage{microtype}      
\usepackage{xcolor}         
\usepackage{amsmath}
\usepackage{algorithm2e}
\usepackage{graphicx}
\usepackage{subfigure}

\usepackage{wrapfig}
\RestyleAlgo{ruled}
\SetKwInput{KwData}{Input}
\SetKwInput{KwResult}{Output}

\SetCommentSty{mycommfont}
\title{Minimizing Energy Costs in Deep Learning Model Training: The Gaussian Sampling Approach}

\author{Challapalli Phanindra Revanth\\
Indian Institute of Technology Hyderabad \\
\texttt{ai22resch01003@iith.ac.in}
\And
Sumohana S Channappayya \\
Indian Institute of Technology Hyderabad \\
\texttt{sumohana@ee.iith.ac.in} \\
\And
C Krishna Mohan \\
Indian Institute of Technology Hyderabad \\
\texttt{ckm@cse.iith.ac.in} \\
}

\newcommand{\param}{\boldsymbol{\theta}}
\newcommand{\boldx}{\boldsymbol{x}}
\newcommand{\loss}{\mathcal{L}}
\newcommand{\bolde}{\mathbf{e}}

\iclrfinalcopy 
\begin{document}

\maketitle

\begin{abstract}
Computing the loss gradient via backpropagation consumes considerable energy during deep learning (DL) model training. In this paper, we propose a novel approach to efficiently compute DL models' gradients to mitigate the substantial energy overhead associated with backpropagation. Exploiting the over-parameterized nature of DL models and the smoothness of their loss landscapes, we propose a method called {\em GradSamp} for sampling gradient updates from a Gaussian distribution. Specifically, we update model parameters at a given epoch (chosen periodically or randomly) by perturbing the parameters (element-wise) from the previous epoch with Gaussian ``noise''. The parameters of the Gaussian distribution are estimated using the error between the model parameter values from the two previous epochs. {\em GradSamp} not only streamlines gradient computation but also enables skipping entire epochs, thereby enhancing overall efficiency. We rigorously validate our hypothesis across a diverse set of standard and non-standard CNN and transformer-based models, spanning various computer vision tasks such as image classification, object detection, and image segmentation. Additionally, we explore its efficacy in out-of-distribution scenarios such as Domain Adaptation (DA), Domain Generalization (DG), and decentralized settings like Federated Learning (FL). Our experimental results affirm the effectiveness of {\em GradSamp} in achieving notable energy savings without compromising performance, underscoring its versatility and potential impact in practical DL applications.
\end{abstract}

\section{Introduction} \label{sec:intro}
Training deep learning (DL) models is infamous for its energy consumption. This aspect has been discussed and debated due to its impact on global warming and sustainable development. The energy consumption of training and hyperparameter tuning of large DL models was studied in (\cite{Strubell_Ganesh_McCallum_2020}). This work recommends AI researchers prioritise the development of energy-efficient hardware and algorithms. A more recent work discusses the energy consumption during inference (\cite{energy}) and suggests that the energy consumed for inference is growing slower than model training. However, this work also cautions that the energy requirements could escalate if the assumptions made become invalidated by increased penetration of AI-based solutions.

Gradient descent-based iterative methods are employed for training DL models. Central to this training process is the forward propagation of data points to compute model predictions, followed by the back-propagation (backprop)  (\cite{backprop}) of the model errors for estimating loss gradients. These gradients are used to update the model parameters. The backprop of model errors is the most energy-consuming operation during training. 

In this work, we focus on reducing the number of backprop operations and improving the energy efficiency of DL model training. Also, the problem of improved energy efficiency in the federated learning (FL) framework  (\cite{fed_intro}) is studied using the same lens.
The contributions of this work include: 
\begin{itemize}
    \item a hypothesis that the gradient distribution of DL models is Gaussian and its empirical evidence,
    \item a simple yet effective strategy called {\em GradSamp} for saving energy during training based on the above hypothesis, and an extensive validation of the strategy using popular DL models on several computer vision tasks,
    \item a stochastic version of federated learning algorithms  (\cite{fedavg,fedprox}) that reduces training rounds, again based on the above hypothesis.
\end{itemize}

\section{Related Works}\label{sec:related}
Given the rise of foundation models  (\cite{liu2023summary}) with massive energy requirements for training, the need for energy efficiency in model training cannot be overemphasized. The design of energy-efficient methods for training DL models has received significant attention from the hardware perspective. An excellent albeit slightly dated survey of efficient processing methods for DL models is presented in  (\cite{Sze2017EfficientPO}).
\cite{esser2015backpropagation} proposed an approach for deploying the backprop algorithm in neuromorphic computing design. This is achieved by treating the spikes and discrete synapses as continuous probability distributions and thereby satisfying the requirement of the backprop algorithm. This approach led to significant energy savings on the TrueNorth neuromorphic architecture  (\cite{merolla2014million}).
Weight pruning is a popular algorithmic approach to efficient model training without compromising performance. \cite{hoefler2021sparsity} present a comprehensive survey of the various techniques proposed in the literature for sparsifying DL models. Recent works (\cite{KUO2023103685, lin2022geometrical}) propose green learning by avoiding non-linearities and the backprop algorithm altogether. MeZO  (\cite{mezo}) proposes the addition of small Gaussian noise to perturb the parameter updates. It requires only two forward passes through the model to compute the gradient estimate.

Our work differs from these approaches in that it neither prunes model parameters nor eliminates the use of the backprop algorithm. Instead, it adopts a stochastic approach to predicting gradient values and reduces the number of forward and backprop operations to achieve energy efficiency. Further, unlike MeZO  (\cite{mezo}), the proposed approach can be used for training DL models from scratch (evaluated here in the sub-LLM parameter count setting).

In the FL framework, a central server coordinates the training of clients by aggregating client model parameters  (\cite{fed_intro}). Each client updates its model parameters locally by training on its private dataset. The clients then share their model parameters with the central server. The server collates the client parameters and computes a set of global weights. These weights are then sent to the clients for use in the next training round. The two-way exchange of model parameters happens over several rounds until the models converge. Let $\boldsymbol{\theta}^{(r)}$ denote the collated parameter vector at the server, and $\boldsymbol{\theta}_k^{(r+1)}$ represent the client $k$'s parameter vector after the client update at round $r$ (using $\boldsymbol{\theta}^{(r)}$ as the initialization). In the FL algorithm, the server update is given by:
\begin{equation}
    \boldsymbol{\theta}^{(r+1)} = \frac{|D_k|}{|D|}\sum\limits_{k=1}^K\boldsymbol{\theta}_k^{(r+1)},
    \label{eqn:fedavg}
\end{equation}
where  $K$ is the number of clients used, $|D_k|$ is the size of data in client $k$ and $|D|$ is the size of overall dataset ($|D|= \sum\limits_{k=1}^{K} |D_k|$). We demonstrate the utility of our hypothesis in the FL framework by reducing the number of communication rounds used to train the clients. 

\section{Proposed Approach}\label{sec:proposed}
In our approach, we concentrate on the classification setting, with the potential for broader application. Let $\mathcal{D} = \{(\boldx_i,y_i)\}_{i=1}^{n}$ represent a dataset comprising of n labeled i.i.d. samples drawn from an unknown but fixed probability distribution $p(\boldx, y)$, where each sample $\boldx_i \in \mathcal{X}$ and $y_i \in \mathcal{Y}$. Our goal is to predict a deep learning predictor $f(\boldsymbol{\theta}): \mathcal{X} \rightarrow \mathcal{Y}$, parameterized by $\param$, with $\mathcal{X}$ as the input space and $\mathcal{Y}$ as the label space. The loss function $\loss(\param)$ can be defined as follows.
\begin{equation}
\loss(\param) = \frac{1}{n}\sum_{i=1}^{n}d(y_i, f(\boldx_i; \param)),    
\label{equn:loss}
\end{equation}
where $d(.,.)$ represents the appropriate distance function measuring the similarity between the predicted and true values.
\par Further considering the standard  Gradient Descent (GD) method where the parameter update at the current epoch $k$ is defined as
\begin{equation} \label{equn:gd}
\begin{split}
    \param^{(k)} &= \param^{(k-1)}-\eta \nabla \loss(\param^{(k-1)}),  \\
  \bolde^{(k)} &= \param^{(k)} - \param^{(k-1)} =  -\eta \nabla \loss(\param^{(k-1)}),
\end{split}
\end{equation}
where  $\mathbf{e}^{(k)}$ denote the error associated with successive parameter updates computed at epochs $k$ and $k-1$. Here, $\eta$ represents the learning rate, and $\nabla\loss(\param)$ denotes the gradient of the loss function (with respect to the model parameters).

\subsection{Hypothesis}\label{sub:hypo}
In this setting, we hypothesize that 
\begin{equation}
  p(\param^{(k)}_i|\param^{(k-1)}_i)\sim \mathcal{N}(\mu_i, \sigma_i^2)
  \label{equn:hypo}
\end{equation}
where $\mu_i$ and $\sigma_i^2$ represent the mean and variance, respectively, of the $i^{th}$ element across the parameter vector $\param$. From equations~\ref{equn:loss} and ~\ref{equn:gd}, the error corresponding to the $\param_i$ be stated as:
\begin{align}\label{equn:error}
    \mathbf{e}_i^{(k)} &= \param_i^{(k)} - \param_i^{(k-1)} \notag \\
    &= -\eta\nabla\left(\frac{1}{n}\sum\limits_{i=1}^n d(y_i, f(\textbf{x}_i;\boldsymbol{\theta}^{(k-1)}))\right) \notag \\
    &= -\eta\left(\frac{1}{n}\sum\limits_{i=1}^n \nabla(d(y_i, f(\textbf{x}_i;\boldsymbol{\theta}^{(k-1)})))\right)
\end{align}
We present supporting arguments for our hypothesis.
\begin{itemize}
    \item Under the assumption of i.i.d. data samples, the central limit theorem (CLT) for large $n$ can be applied to the average gradient computed over the training set (shown in the right hand side of (\ref{equn:error}))  (\cite{pmlr-v202-francazi23a}).   
    \item The next argument is that DL models are over-parameterized, and that the loss landscape of such models is smooth  (\cite{liu2022loss}). This is often characterized using the $K$-Lipschitz smoothness condition expressed as $||\mathcal{L}(\boldsymbol{\theta}^{(k)}) - \mathcal{L}(\boldsymbol{\theta}^{(k-1)}) || \leq K ||\boldsymbol{\theta}^{(k)} - \boldsymbol{\theta}^{(k-1)} ||$. This means that the parameters updates evolve slowly over iterations. Based on this smoothness condition, we infer that error between successive updates is usually small and close to zero. This translates to the error distribution being unimodal and centered around zero.  
     \item If the i.i.d. assumption is restrictive and unrealistic, Chen and Shao  (\cite{chen2004normal}) present CLT-type results for $m-$dependent zero-mean random fields with bounded second and third moments. Given the smoothness condition of the loss landscape, it is reasonable to assume that the gradient elements have zero mean and their second and third order moments are bounded. This result provides further justification for our hypothesis.
\end{itemize}
In practical applications, Gaussian distribution parameters are typically determined via fitted likelihood estimation over the error vector $\bolde^{(k)}$. However, due to the large parameter count in deep learning models, we introduce a simplified method to compute the mean and variance across each element of $\bolde^{(k)}$, a technique we will elaborate on shortly.
\subsection{Energy-Efficient Model Training}\label{sub:energy}
Since DL models are over-parameterized we proposed a simple approach to estimate the parameters of Gaussian distribution. Consider that the model consists of $L$ layers, the error vector $\bolde^{(k)}$ is partitioned into 
$\mathbf{e}^{(k)} = [\mathbf{e}^{(k)}_1| \mathbf{e}^{(k)}_2| \cdots| \mathbf{e}^{(k)}_L]$. Here, $\bolde^{(k)}_l$ denotes the error sub-vector corresponding to epoch $k$ for layer $l$, computed using Equation ~\ref{equn:error}. We further assume that each element in $\bolde^{(k)}_l$ follows an i.i.d. distribution $\mathcal{N}(\mu_l^{(k)}, (\sigma_l^{(k)})^2)$, where $\mu_l^{(k)}$ and $(\sigma_l^{(k)})^2$ represent the mean and variance, respectively. These parameters are estimated via fitted likelihood estimation for each layer $l$. This assumption is supported by the similarity in dynamic ranges between the inputs and outputs at a given layer, attributed to operations like batch normalization and layer normalization. While this may seem to oversimplify the scenario, it has demonstrated effectiveness in practical applications.\\
Now apply our hypothesis to achieve our energy-efficient training. 
By updating the parameter update value at $k+1$ epoch for each layer $l$ as:
\begin{equation}\label{equn:energy}
  \param^{(k+1)}_l = \param^{(k)}_l + \mathbf{\tilde{e}}^{(k)}_l  \hspace{10mm} \forall l=1,2,\dots, L,
\end{equation}
where $ \Tilde{\bolde}^{(k)}_l $ represents the sampled elements drawn from the distribution $ \mathcal{N}(\mu_l^{(k)}, (\sigma_l^{(k)})^2) $, with parameters obtained through fitted likelihood estimation. Equation ~\ref{equn:energy} elucidates our approach of skipping the entire $k+1$ epoch by leveraging the parameters and fitted likelihood from previous error updates. This approach which we call {\em GradSamp} is illustrated in Algorithm ~\ref{algo:sto}.\
One natural question arises about the frequency of our sampling strategy. To delve deeper, we explore different sampling strategies in Section ~\ref{sub:strategy}.

\begin{algorithm}[H]
 \KwData{Model parameters $\boldsymbol{\theta}$, Training Dataset $\mathcal{D}=\{(x_k, y_k)\}_{k=1}^{n}$, Layer count  $L$, number of parameters at each layer $\{n_l\}_{l=1}^{L}$, $\epsilon = 0.001$}
 \KwResult{Trained parameters $\boldsymbol{\theta^*}$}
 Initialize: $k \leftarrow 0$ \tcp*{epoch count}
 \While{Stopping Condition Not Met}{
 \eIf {Sampling Condition Met} {
 Compute error at $(k-1)^{th}$ epoch~\\
 $\bolde^{(k-1)} = \param^{(k-1)} - \param^{(k-2)}$ \tcp*{From buffer}
 $\bolde^{(k-1)}$ is partitioned layer-wise~\\
 $\bolde^{(k-1)} = [\bolde^{(k-1)}_1|\bolde^{(k-1)}_2|\ldots|\bolde^{(k-1)}_L ]$~\\
 \For{$l\gets1$ \KwTo $L$}{
    $\mu_l^{(k-1)} = \frac{1}{n_l}\sum_{i=1}^{n_l}\bolde^{(k-1)}_l[i]$~\\
    $(\sigma_l^{(k-1)})^2 = \frac{1}{n_l}\sum_{i=1}^{n_l}(\bolde^{(k-1)}_l[i] - \mu_l^{(k-1)})^2 + \epsilon $~\\

    $\mathbf{\Tilde{e}}_l^{(k-1)} \sim \mathcal{N}(\mu_l^{(k-1)}, (\sigma_l^{(k-1})^2) \hspace{5mm} (n_l \text{ times})$~\\  
    $\param_l^{(k)} = \param_l^{(k-1)} + \mathbf{\Tilde{e}}_l^{(k-1)}$
    }  
 }
 {
 Save $\boldsymbol{\theta}^{(k-1)}$ in buffer~\\
 $\boldsymbol{\theta}^{(k)} = \boldsymbol{\theta}^{(k-1)} - \eta \nabla\mathcal{L}(\boldsymbol{\theta}^{(k-1)})$ \tcp*{Update using backprop}
 }
 $k \leftarrow k+1$ \tcp*{Update epoch count}
 }
 \caption{The {\em GradSamp} Algorithm}
 \label{algo:sto}
\end{algorithm}
\subsection{Stochastic Federated Learning Algorithms}
\label{ssec:fedavg} 
We found the FL framework to be a natural fit to test our hypothesis in a distributed learning setting. Specifically, we claim that the elements of $\mathbf{e}^{(r+1)} = \boldsymbol{\theta}^{(r+1)} - \boldsymbol{\theta}^{(r)}$ follow a unimodal Gaussian distribution (using the notation from (\ref{eqn:fedavg}), and $r$ being the round index). In other words,
\begin{equation}
    p(\boldsymbol{\theta}_i^{(r+1)}| \boldsymbol{\theta}_i^{(r)})\sim \mathcal{N}(\mu_i^{(r)}, (\sigma_i^{(r)})^2)
\end{equation}
We modify the {\tt FedAvg}  (\cite{fedavg}) and {\tt{FedProx}}  (\cite{fedprox}) algorithms to their stochastic variants where the update rule becomes:
\begin{equation}
    \boldsymbol{\theta}_i^{(r+1)} = \boldsymbol{\theta}_i^{(r)} + \tilde{\bolde}_i^{(r+1)},
\end{equation}
where $\mathbf{\tilde{e}}_i^{(r+1)}$ corresponds to samples drawn from the distribution $\mathcal{N}(\mu_i^{(r)}, (\sigma_i^{(r)})^2)$. As before, we apply the layer-level simplification to the FL setting as well. A periodic sampling strategy is used for skipping communication rounds in this setting. We experiment with the sampling period to assess its impact on performance.
\section{Experiments and Results}
We conducted experiments across various computer vision tasks, including image classification, object detection, and semantic segmentation. We also evaluated our hypothesis in out-of-distribution tasks like Domain Adaptation (DA) and Domain Generalization (DG), and explored its application in Federated Learning (FL) strategies. For image classification, we assessed the efficacy of ResNet-50 (\cite{Resnet}), Swin-Transformer (\cite{swin}), and MLP-Mixer (\cite{mlp-mixer}). For the context of object detection, we experimented with YOLOv7 (\cite{yolov7}) and RT-DETR (\cite{rt-detr}). For semantic segmentation, we explored DeepLabv3 (\cite{Deeplabv3}), U-Net (\cite{unet}), and SegFormer (\cite{segformer}). In Domain Adaptation (DA), we experimented with MDD (\cite{MDD}) and MCC (\cite{MCC}), while in Domain Generalization (DG), we evaluated VREx (\cite{VREx}) and GroupDRO (\cite{GroupDRO}). Finally, for Federated Learning (FL), we tested standard algorithms like {\tt FedAvg}  (\cite{fedavg}) and {\tt FedProx} (\cite{fedprox}) (with a proximal parameter $\mu = 0.2$) on lightweight models such as MobileNetV2  (\cite{mobilenetv2}), ShuffleNetV2  (\cite{shufflenetv2}), SqueezeNet  (\cite{squeezenet}), and ResNet-50  (\cite{Resnet}). Implementation details are provided in section~\ref{sec:sup_expdetails} of the supplementary material.
\subsection{Datasets and Evaluation metrics}\label{sub:dataset}
For image classification, we utilized standard datasets including CIFAR-10  (\cite{cifar10}), CIFAR-100  (\cite{cifar10}) and Tiny ImageNet  (\cite{tiny}). For object detection, we conducted experiments on the Pascal-VOC-2012  (\cite{pascal12}) dataset. We employed standard COCO format metrics such as {\it mAP$_{.5}$} and  {\it mAP$_{.5: .95}$} as the mean of {\it APs$@[.5, .5: .95]$} respectively. In the case of image segmentation, We experimented with the CityScapes  (\cite{cityscapes}) dataset and used the Jaccard Index as a metric, providing the mean Intersection over Union (mIOU) for our evaluations.. In our Domain Adaptation (DA) experiments, we utilized the Office-Home  (\cite{office31}) and Vis-DA  (\cite{visda}) datasets. For Domain Generalization (DG), we conducted experiments using the PACS  (\cite{pacs}) and Office-Home  (\cite{officehome}) datasets. Finally, for Federated Learning (FL) setting, we conducted experiments with CIFAR-10  (\cite{cifar10} and CIFAR-100  (\cite{cifar10}) datasets toward the classification setting. We relied on the standard evaluation metric of accuracy {\it Acc} (\%) for all classification tasks.

\subsection{Sampling Strategies}\label{sub:strategy}
Algorithm ~\ref{algo:sto} naturally prompts an inquiry into the optimal frequency or interval of sampling. Several strategies could be employed to answer this question. A simple strategy is to make this update periodic and experiment with the period. Another approach is to randomly choose an epoch and apply (\ref{equn:energy}). Yet another strategy is to rely on the validation loss and enable (\ref{equn:energy}) once this loss drops below a certain threshold. 
In the periodic case, we experimented with periods (Pe) of 5 and 10 epochs. In the case of probabilistic sampling (Pr), we considered three different probabilities: 0.2, 0.5 and 0.7. In other words, we drew samples from a {\tt Bernoulli(p)} distribution with $p = 0.2, 0.5 \text{ and } 0.7$. The gradient is sampled from the Gaussian distribution every time we draw a 1. In the case of Delayed Periodic (DP) sampling, we initiate the process by completing the first half of the epochs without sampling and start periodic sampling with periods of 5 and 10 in the latter half. Similarly, for Delayed Random (DR) sampling, we execute half of the epochs using the standard approach and then initiate random sampling with probabilities of 0.2, 0.5, and 0.7.
\subsection{Results}

\begin{figure*}[!ht]
  \subfigure[ResNet-50, 1$^{\text{st}}$ $Conv$ layer \\of 2$^{\text{nd}}$ stage, 1$^{\text{st}}$ Residual block, \\Epoch 50]{
    \includegraphics[width=0.3\textwidth]{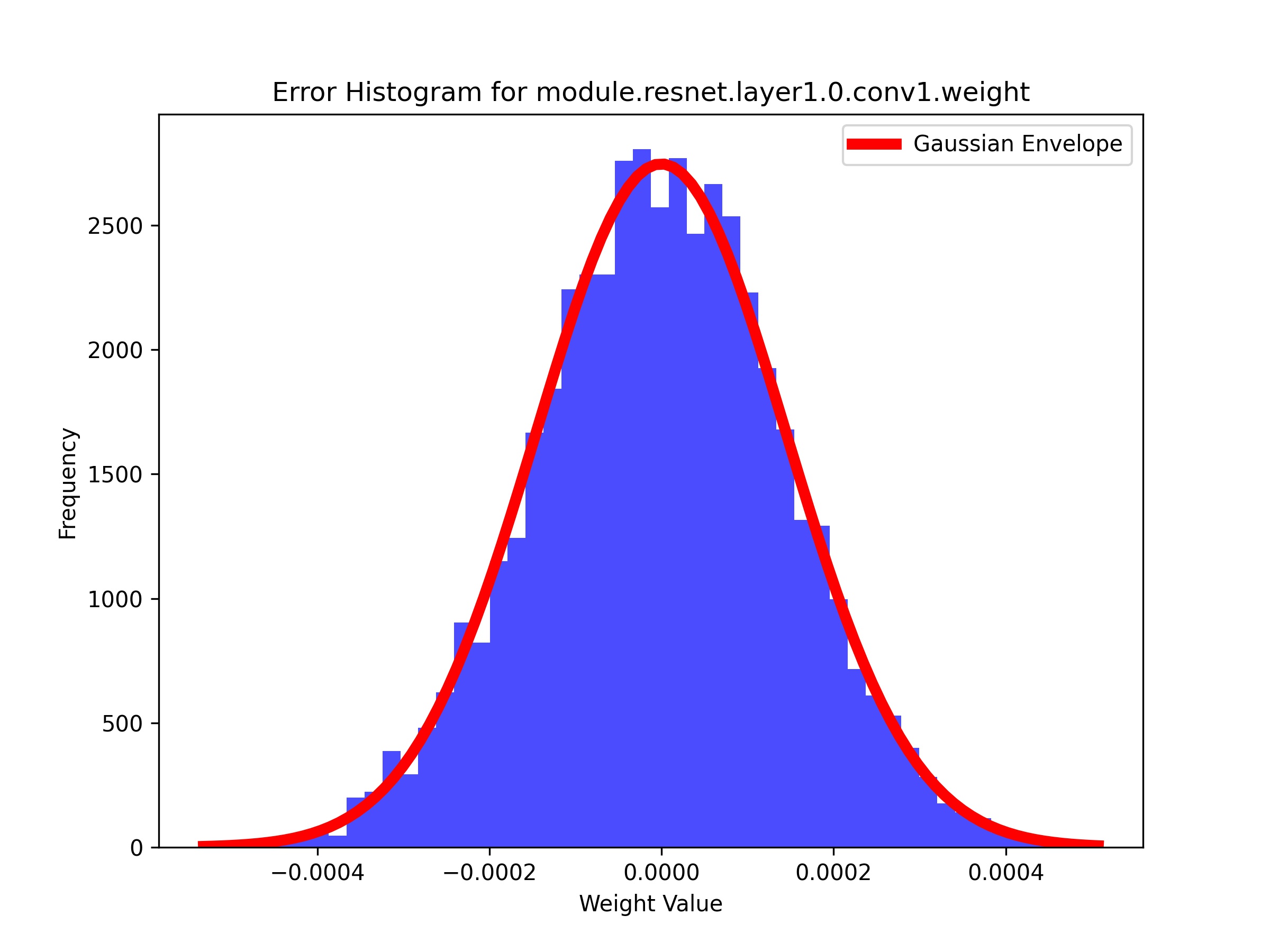}
  }
  \subfigure[ResNet-50, 1$^{\text{st}}$ $Conv$ layer \\of 2$^{\text{nd}}$ stage, 1$^{\text{st}}$ Residual block, \\Epoch 150]{
    \includegraphics[width=0.3\textwidth]{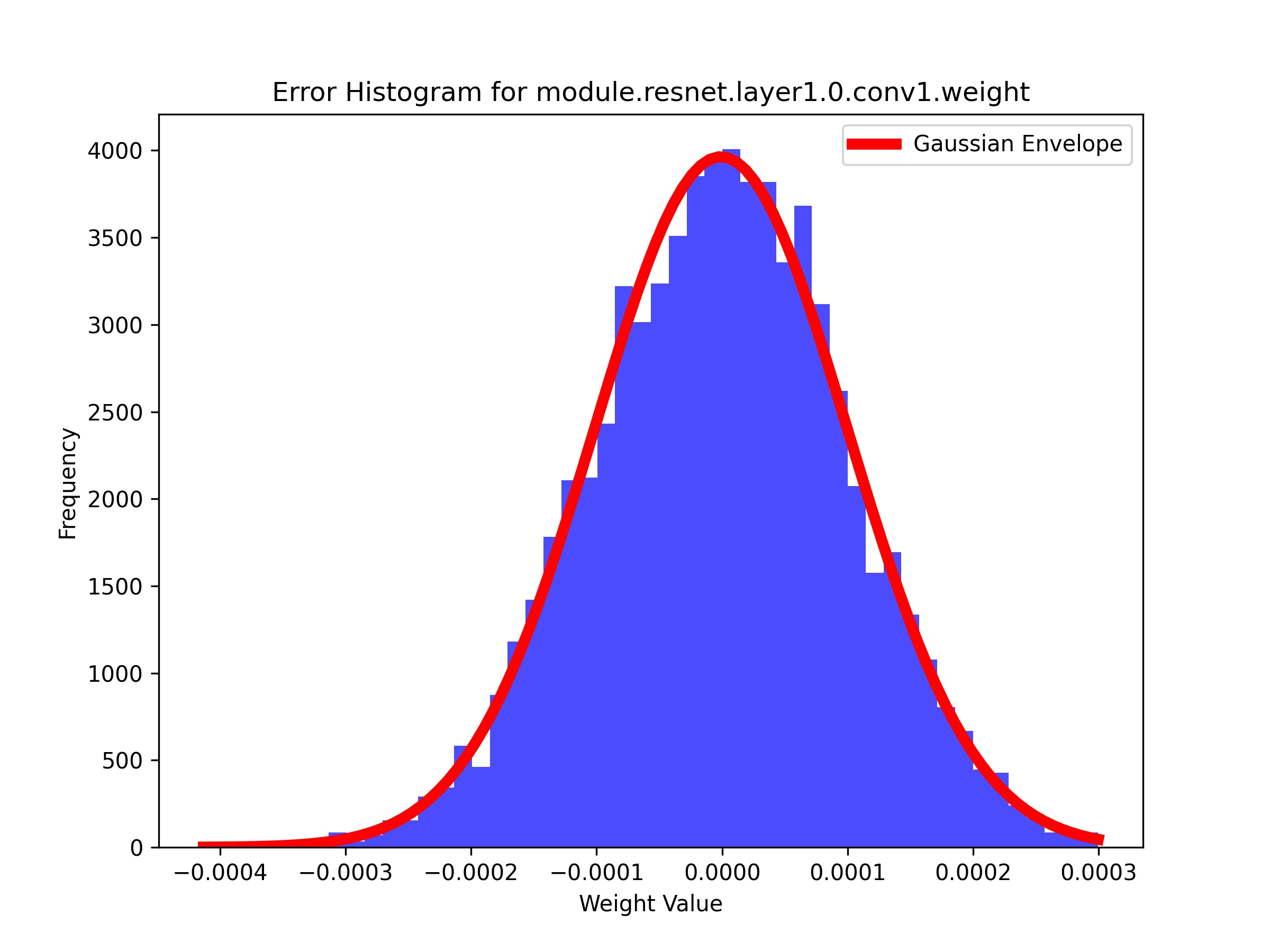}
  }
  \subfigure[ResNet-50, 1$^{\text{st}}$ $Conv$ layer \\of 2$^{\text{nd}}$ stage, 1$^{\text{st}}$ Residual block, \\Epoch 200]{
    \includegraphics[width=0.3\textwidth]{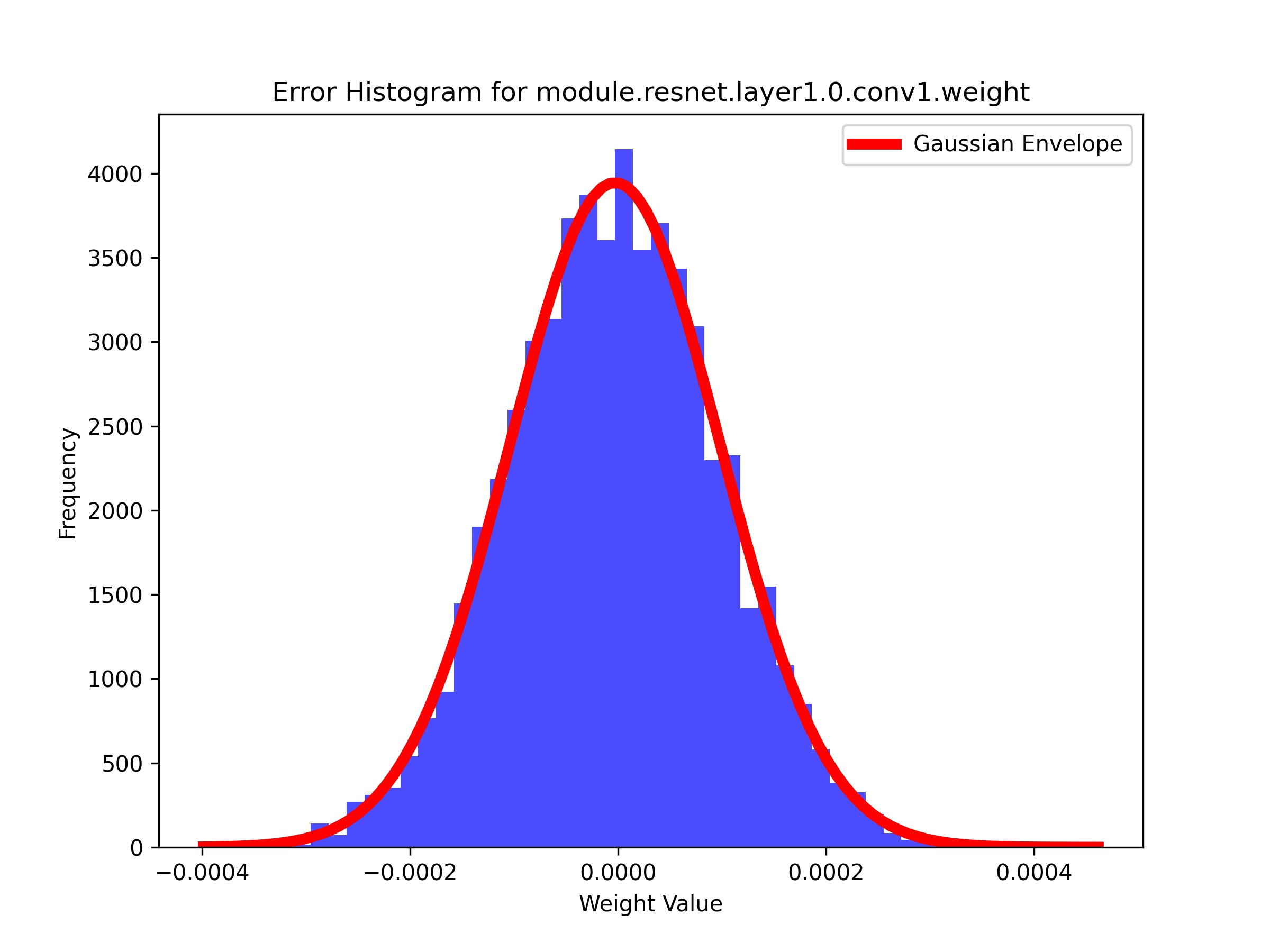}
  }
  \caption{ResNet-50 error histograms plotted at different epochs, with gradients sampled every $10$ epochs on the CIFAR-10 dataset.}
  \label{fig:hypo_resnet}
\end{figure*}
\begin{figure*}[!ht]
  \subfigure[Swin Transformer, $QKV$ weights corresponds to \\ 2$^{\text{nd}}$ Residual block, 2$^{\text{nd}}$ stage, \\ 1$^{\text{st}}$ block, 2$^{\text{nd}}$ sub-component, \\2$^{\text{nd}}$ layer, Epoch 50]{
    \includegraphics[width=0.3\textwidth]{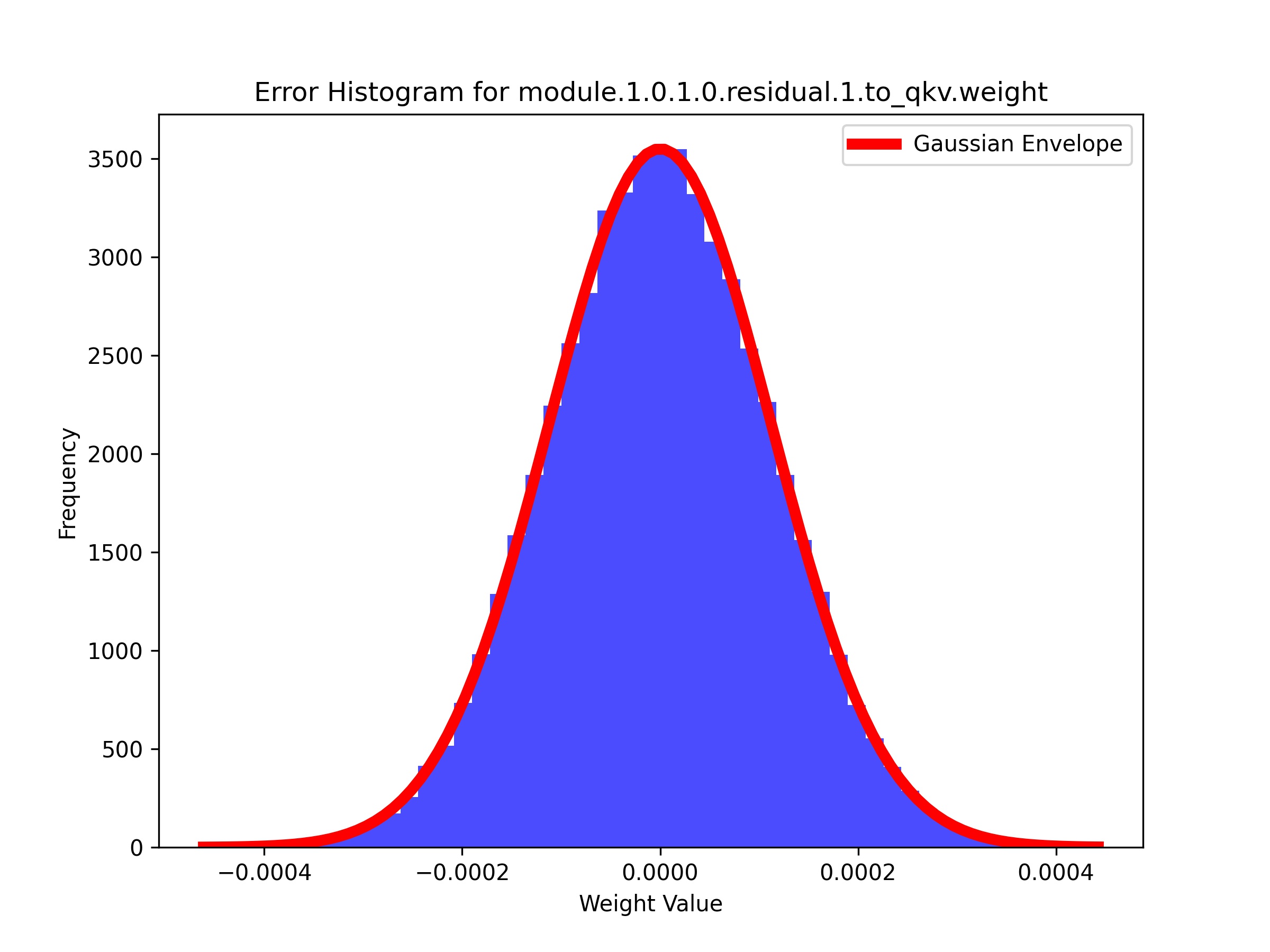}
  }
  \subfigure[Swin Transformer, $QKV$ weights corresponds to \\ 2$^{\text{nd}}$ Residual block, 2$^{\text{nd}}$ stage, \\ 1$^{\text{st}}$ block, 2$^{\text{nd}}$ sub-component, \\2$^{\text{nd}}$ layer, Epoch 100]{
    \includegraphics[width=0.3\textwidth]{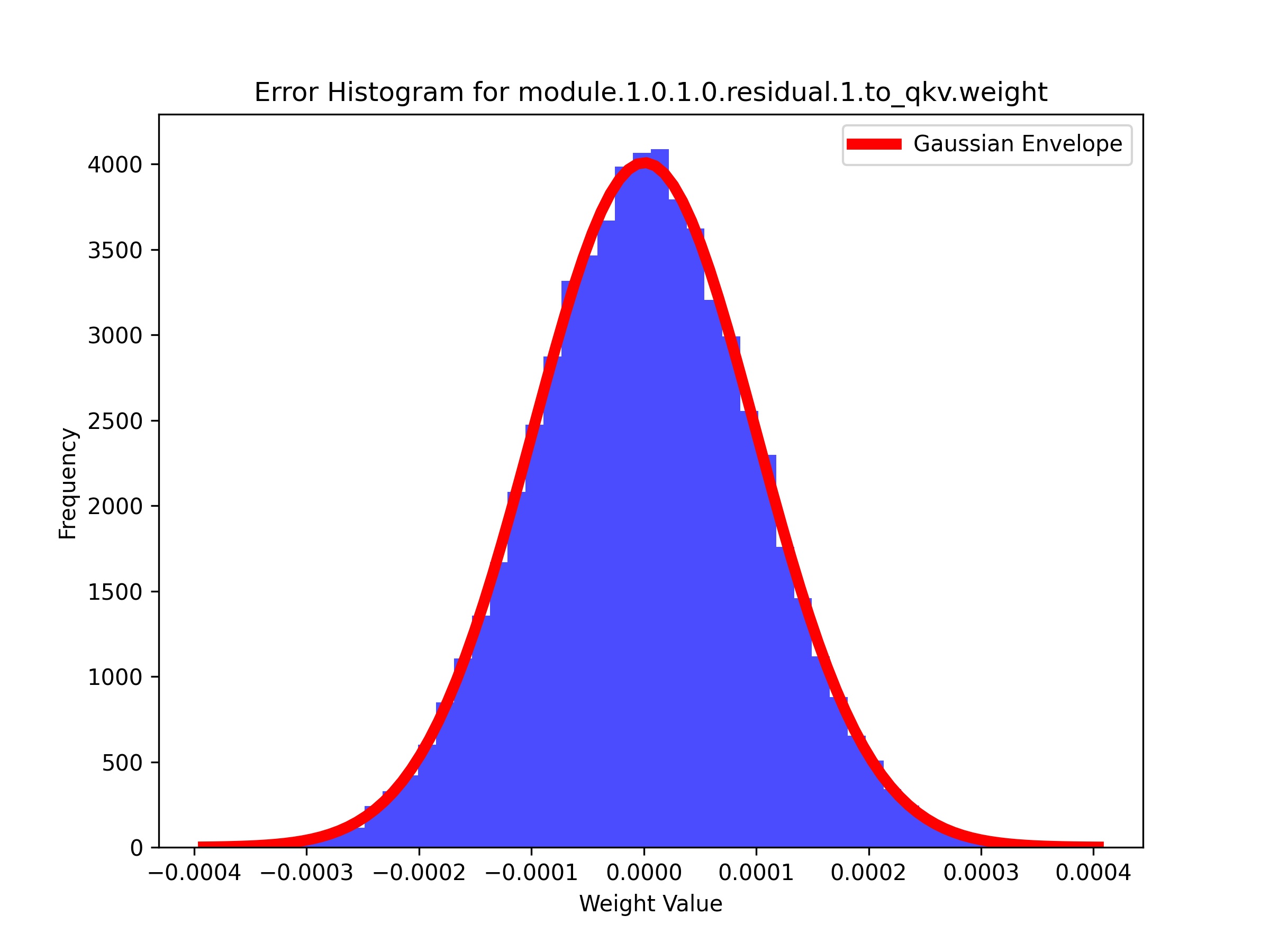}
  }
  \subfigure[Swin Transformer, $QKV$ weights corresponds to \\ 2$^{\text{nd}}$ Residual block, 2$^{\text{nd}}$ stage, \\ 1$^{\text{st}}$ block, 2$^{\text{nd}}$ sub-component, \\2$^{\text{nd}}$ layer, Epoch 150]{
    \includegraphics[width=0.3\textwidth]{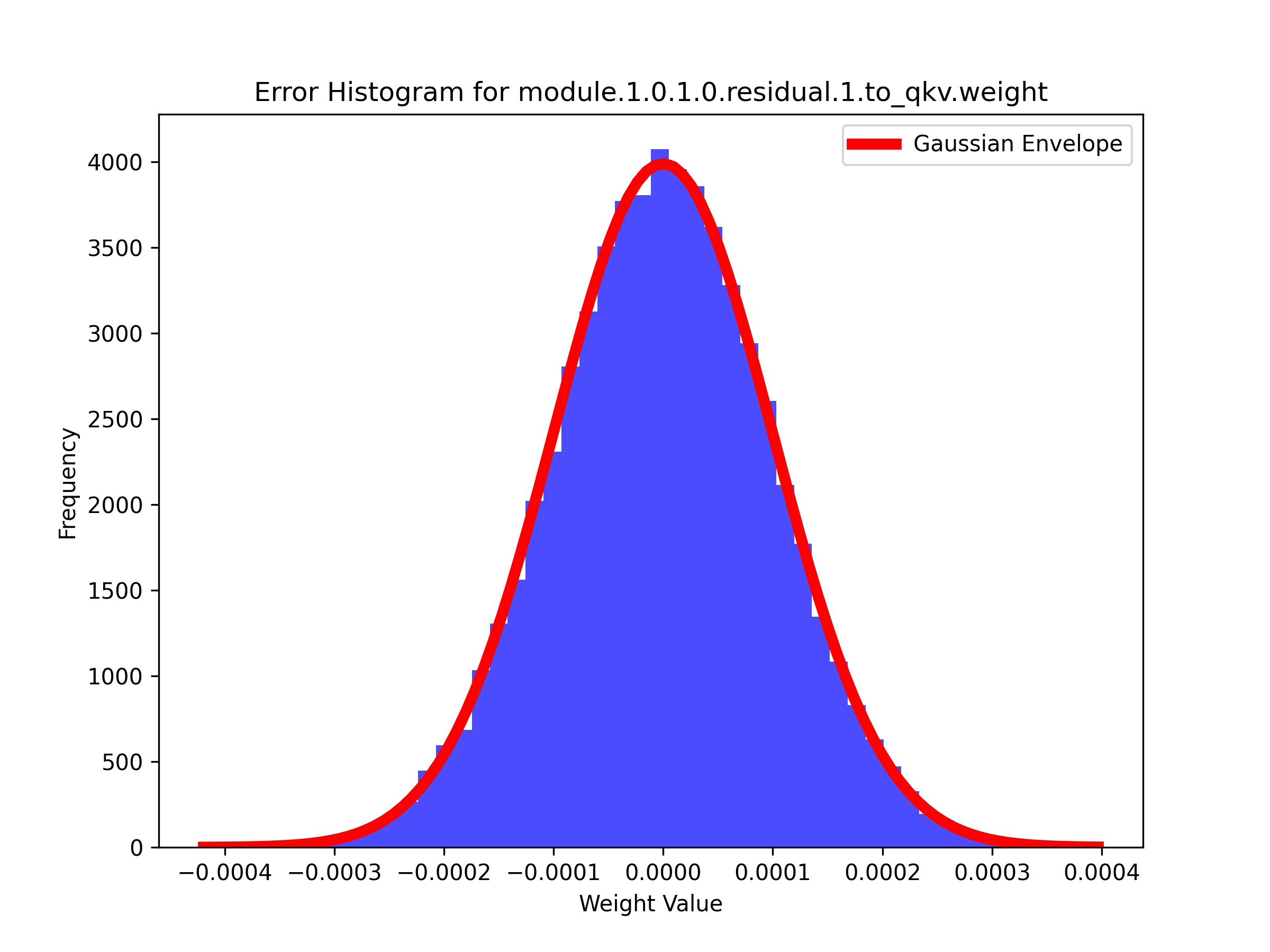}
  }
  \caption{Swin Transformer error histograms plotted at different epochs, with gradients sampled every $10$ epochs on the CIFAR-10 dataset.}
  \label{fig:hypo_swin}
\end{figure*}
\begin{figure*}[!ht]
  \subfigure[MLP-Mixer, 1$^{\text{st}}$ $FC$ layer, \\3$^{\text{rd}}$ block, 2$^{\text{nd}}$ stage, Epoch 50]{
    \includegraphics[width=0.3\textwidth]{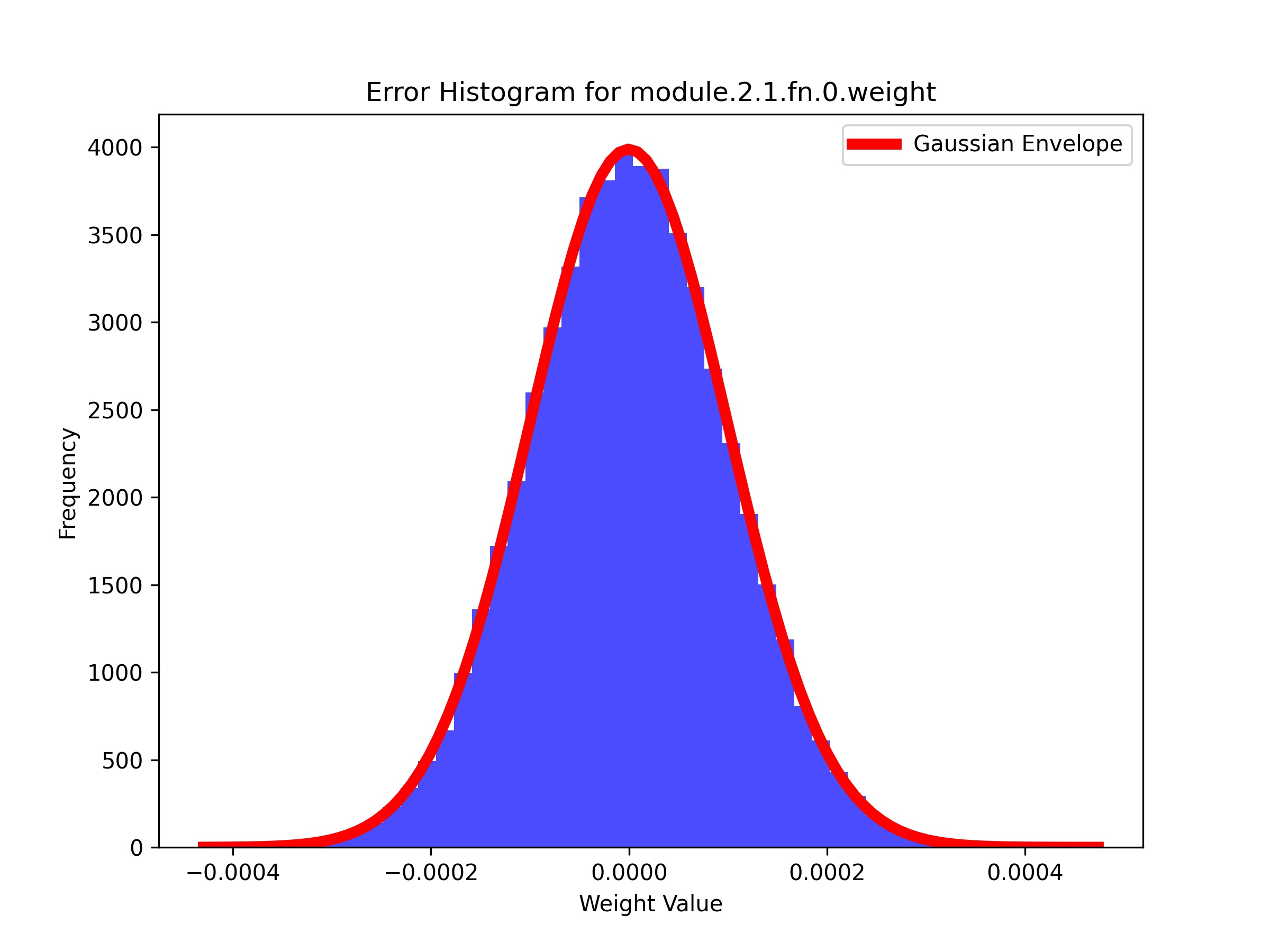}
  }
  \subfigure[MLP-Mixer, 1$^{\text{st}}$ $FC$ layer, \\3$^{\text{rd}}$ block, 2$^{\text{nd}}$ stage, Epoch 150]{
    \includegraphics[width=0.3\textwidth]{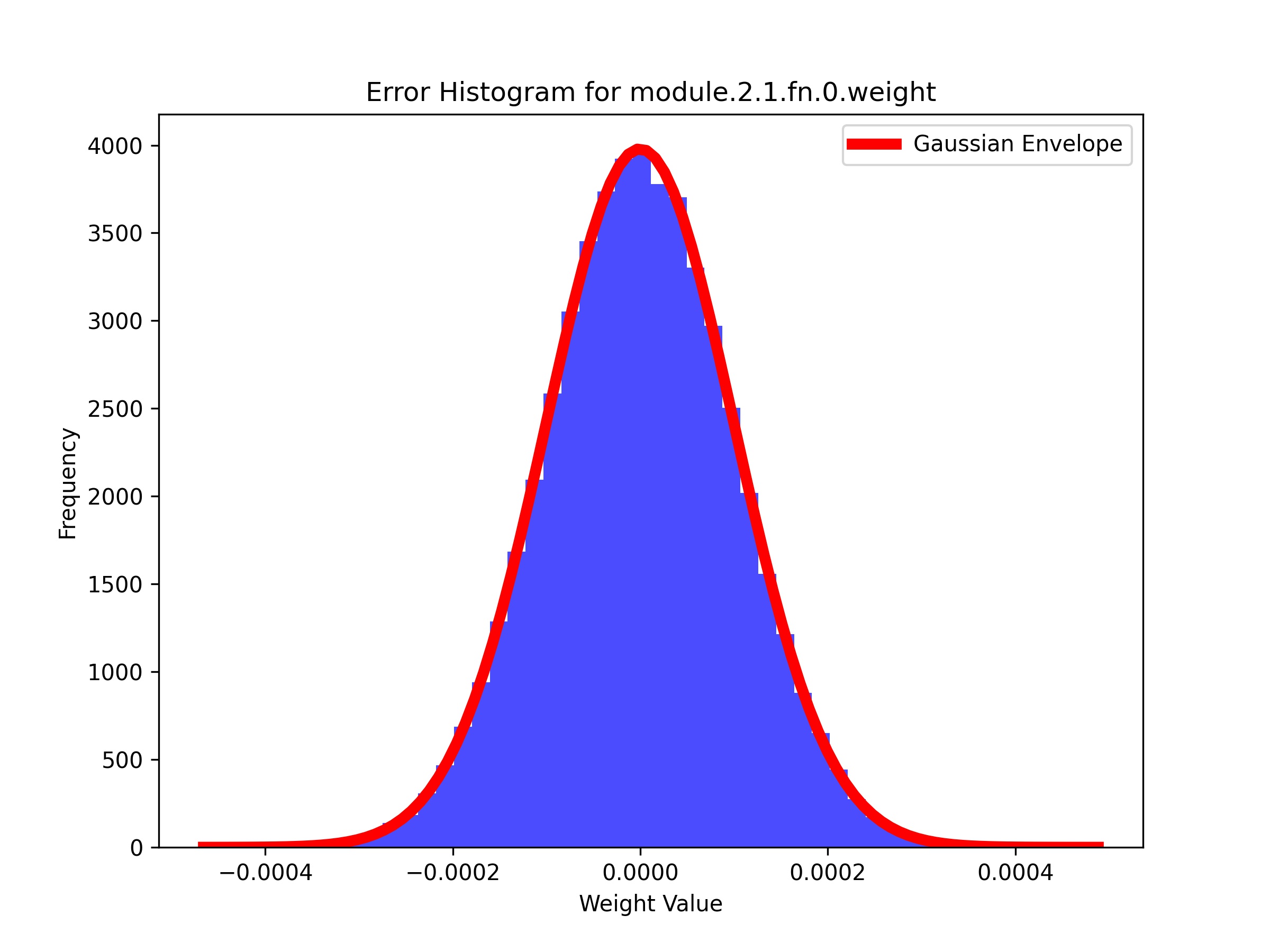}
  }
  \subfigure[MLP-Mixer, 1$^{\text{st}}$ $FC$ layer, \\3$^{\text{rd}}$ block, 2$^{\text{nd}}$ stage, Epoch 200]{
    \includegraphics[width=0.3\textwidth]{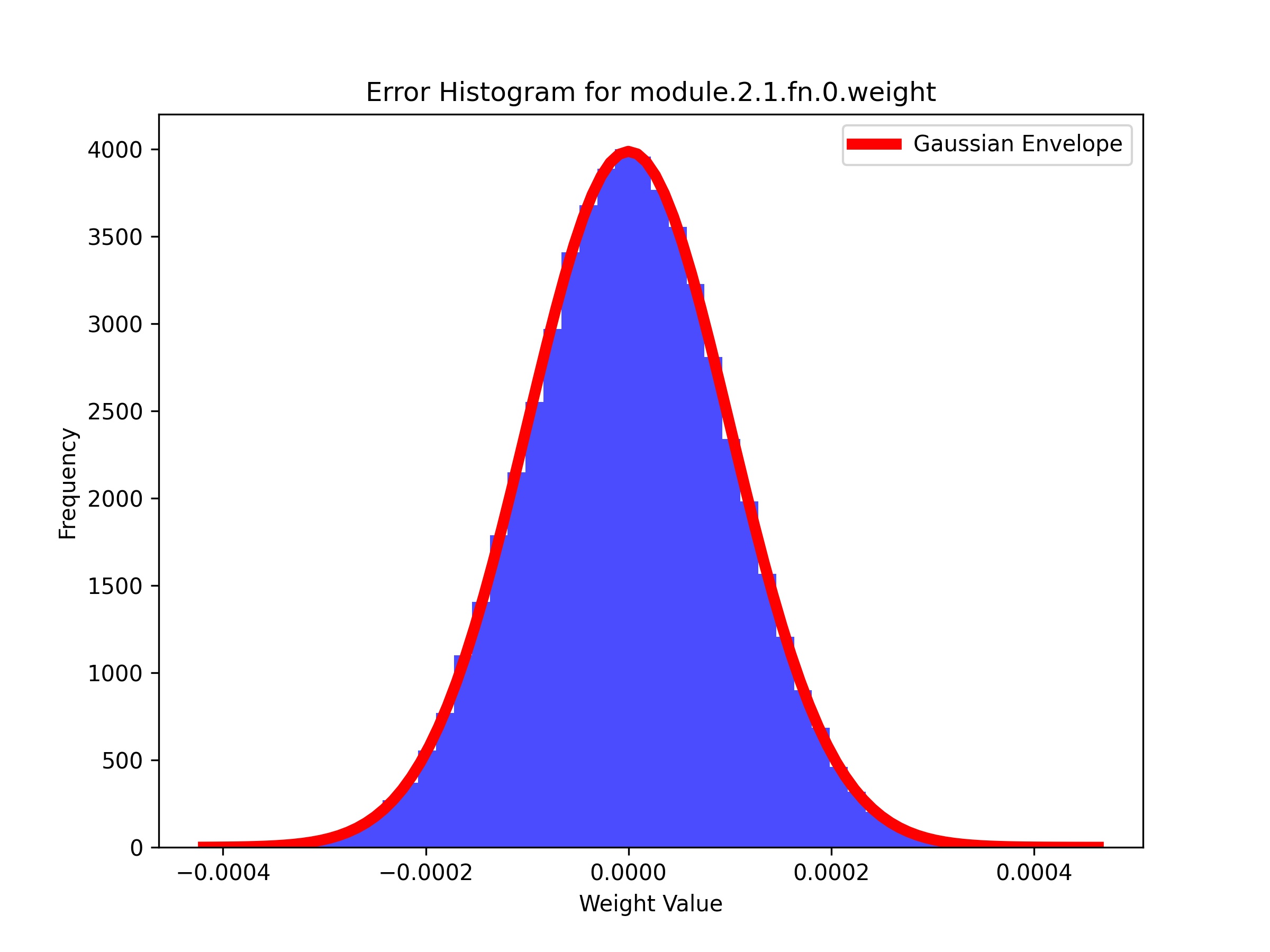}
  }
  \caption{MLP-Mixer error histograms plotted at different epochs, with gradients sampled every $10$ epochs on the CIFAR-10 dataset.}
  \label{fig:hypo_mixer}
\end{figure*}
{\bf Image Classification:} For the image classification task, we evaluate our sampling strategies across different baseline networks  (\cite{Resnet, swin, mlp-mixer}). We first present qualitative results that provide empirical evidence supporting our hypothesis. The error histograms (refer (~\ref{equn:error})) for specific layer(s) over different epochs for ResNet-50  (\cite{Resnet}), Swin Transformer  (\cite{swin}) and MLP-Mixer  (\cite{mlp-mixer}) models, as shown in Figs~\ref{fig:hypo_resnet},~\ref{fig:hypo_swin}, and ~\ref{fig:hypo_mixer}, respectively, illustrate instances where our hypothesis holds for lightweight, standard, and recent transformer-based models. The standard normality test   (\cite{pearson}) with a p-value threshold of 5\% is applied

\begin{figure*}[!ht]
  \subfigure[ResNet-50, 3$^{\text{rd}}$ $Batch \hspace{1mm} Norm$ layer of 5$^{\text{th}}$ stage, 1$^{\text{st}}$ \\Residual block, Epoch 150]{
    \includegraphics[width=0.3\textwidth]{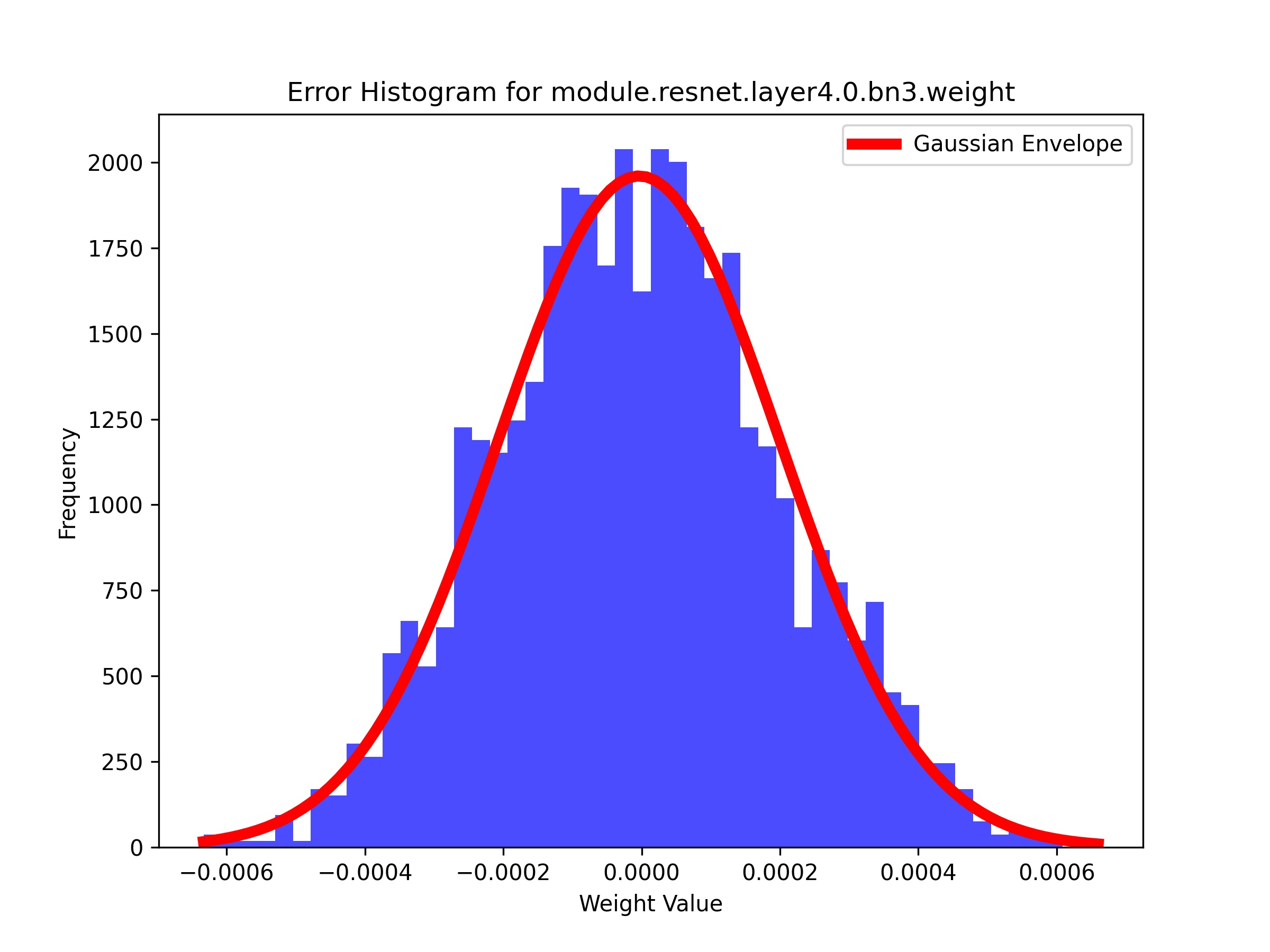}}
  \subfigure[Swin Transformer, 1$^{\text{st}}$ \\ $Residual \hspace{1mm} block$, 2$^{\text{nd}}$ stage, \\ 1$^{\text{st}}$ block, 2$^{\text{nd}}$ sub-component, \\ 1$^{\text{st}}$ layer,  Epoch 140]{
    \includegraphics[width=0.3\textwidth]{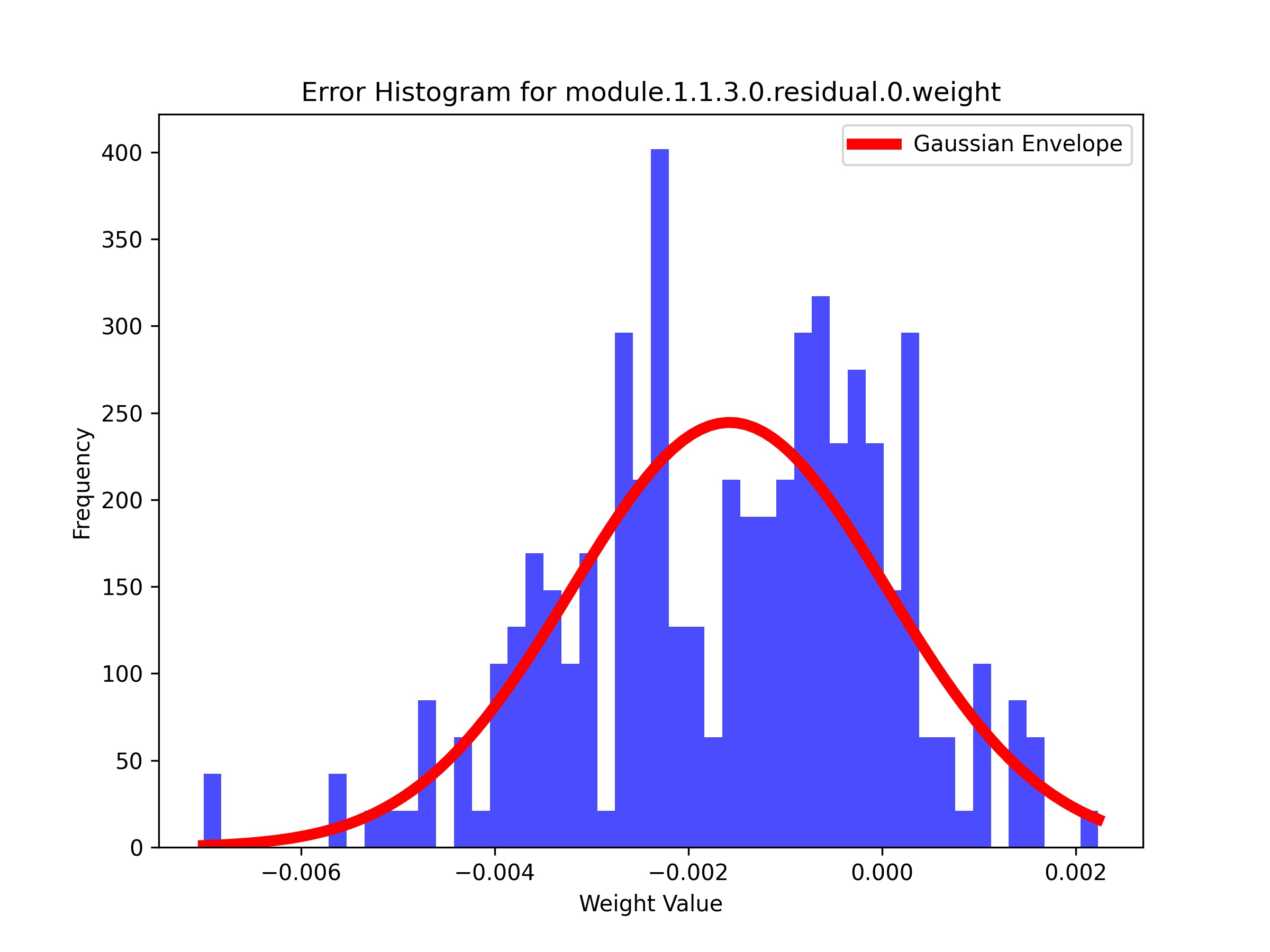}
  }
  \subfigure[MLP-Mixer, $Norm \hspace{1mm} weights$ corresponds to, 3$^{\text{rd}}$ block, 1$^{\text{st}}$ stage, Epoch 200]{
    \includegraphics[width=0.3\textwidth]{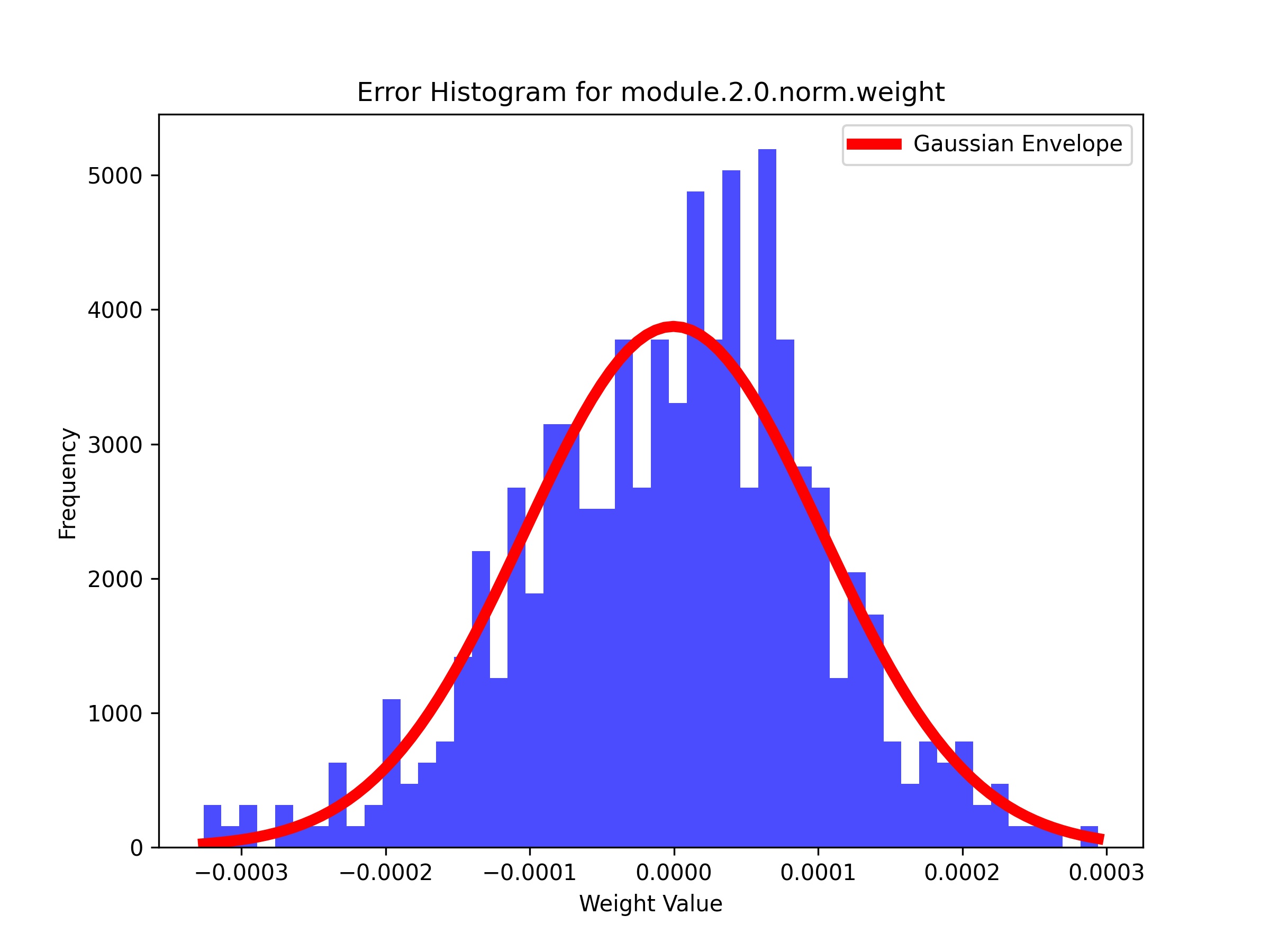}
  }
  \caption{Examples, where our hypothesis fails, include cases where the normality test did not hold. Despite the failure of the normality test, it's noteworthy that the histograms are mostly unimodal and could be modeled using a skewed Gaussian function.}
  \label{fig:hypo_fail}
\end{figure*}
Upon closer examination of the failure cases depicted in plot ~\ref{fig:hypo_fail}, we observe error plots exhibiting characteristics of a skewed unimodal distribution. Despite their deviation from perfect symmetry, these distributions can still be adequately approximated by a unimodal Gaussian distribution. These findings highlight the nuanced nature of error distributions in deep learning models and emphasize the importance of robust statistical analysis in model evaluation and interpretation. Additional experiments are detailed in section~\ref{sec:sstandard} of the supplementary material that provide comprehensive insights.
\begin{table}[!ht]
    \tiny
    \caption{
Performance comparison across various standard and non-standard models under diverse sampling strategies, including periodic (Pe), probabilistic (Pr), Delayed Periodic (DP), and Delayed Random (DR) sampling methods, along with the total TFLOPS (in Tera FLOPS) across 5 different trials. }
    \centering
    \begin{tabular}{cl|ccc}
    \hline
       { \bf Model}  &{ \bf Strategy (Savings \%)}&{ \bf CIFAR-10/ TFLOPS}&{ \bf CIFAR-100/ TFLOPS}&{ \bf TINY/ TFLOPS}   \\
       \hline
\multirow{11}{*}{{\bf ResNet-50  (\cite{Resnet})}}
&Baseline (0)	&80.49	$_{\pm 0.11}$	/ 1685.44	&40.90	$_{\pm 0.56}$	/ 1685.90	&29.29	$_{\pm 0.36}$	/ 13422.23 \\
&Pe= 5 (20)	&80.55	$_{\pm 0.60}$	/ 1407.34	&38.15	$_{\pm 0.44}$	/ 1407.73	&29.10	$_{\pm 0.11}$	/ 11207.56 \\
&Pe= 10 (10)	&79.64	$_{\pm 0.41}$	/ 1542.18	&40.15	$_{\pm 0.20}$	/ 1542.60	&33.12	$_{\pm 0.17}$	/ 12281.34 \\
&Pr= 0.2 (20)	&80.08	$_{\pm 0.10}$	/ 1483.19	&40.12	$_{\pm 0.46}$	/ 1483.59	&30.22	$_{\pm 0.46}$	/ 11811.56 \\
&Pr= 0.5 (50)	&79.13	$_{\pm 0.80}$	/ 1137.67	&39.41	$_{\pm 0.39}$	/ 1137.98	&27.37	$_{\pm 0.34}$	/ 9060.00 \\
&Pr= 0.7 (70)	&75.20	$_{\pm 0.75}$	/ 1002.84	&40.60	$_{\pm 0.22}$	/ 1003.11	&29.04	$_{\pm 0.10}$	/ 7986.23 \\
&DP = 5 (10)	&79.28	$_{\pm 0.34}$	/ 1601.17	&39.63	$_{\pm 0.69}$	/ 1601.61	&28.21	$_{\pm 0.21}$	/ 12751.12 \\
&DP = 10 (5)	&78.38	$_{\pm 0.29}$	/ 1643.30	&40.55	$_{\pm 0.44}$	/ 1643.75	&28.98	$_{\pm 0.08}$	/ 13086.67 \\
&DR = 0.2 (10)	&78.76	$_{\pm 0.35}$	/ 1601.17	&41.13	$_{\pm 0.20}$	/ 1601.61	&28.29	$_{\pm 0.22}$	/ 12751.12 \\
&DR = 0.5 (25)	&78.13	$_{\pm 0.20}$	/ 1474.76	&38.42	$_{\pm 0.32}$	/ 1475.16	&29.41	$_{\pm 0.43}$	/ 11744.45 \\
&DR = 0.7 (35)	&77.11	$_{\pm 0.26}$	/ 1390.49	&36.27	$_{\pm 0.25}$	/ 1390.87	&24.43	$_{\pm 0.24}$	/ 11073.34 \\ \hline
\multirow{11}{*}{\bf Swin  (\cite{swin})}
&Baseline (0)	&80.68	$_{\pm 0.65}$	/ 8261.52	&52.23	$_{\pm 0.55}$	/ 8261.98	&42.12	$_{\pm 0.34}$	/ 66093.79 \\
&Pe= 5 (20)	&79.58	$_{\pm 0.49}$	/ 6898.37	&52.51	$_{\pm 0.47}$	/ 6898.75	&41.39	$_{\pm 0.43}$	/ 55188.32 \\
&Pe= 10 (10)	&79.20	$_{\pm 0.45}$	/ 7559.29	&47.16	$_{\pm 0.19}$	/ 7559.71	&42.49	$_{\pm 0.34}$	/ 60475.82 \\
&Pr= 0.2 (20)	&78.37	$_{\pm 0.40}$	/ 7270.14	&49.42	$_{\pm 0.44}$	/ 7270.54	&39.25	$_{\pm 0.30}$	/ 58162.54 \\
&Pr= 0.5 (50)	&74.33	$_{\pm 0.23}$	/ 5576.53	&48.25	$_{\pm 0.24}$	/ 5576.84	&38.54	$_{\pm 0.15}$	/ 44613.31 \\
&Pr= 0.7 (70)	&71.01	$_{\pm 0.11}$	/ 4915.60	&50.12	$_{\pm 0.25}$	/ 4915.88	&28.23	$_{\pm 0.18}$	/ 39325.81 \\
&DP = 5 (10)	&79.49	$_{\pm 0.16}$	/ 7848.44	&50.43	$_{\pm 0.27}$	/ 7848.88	&37.09	$_{\pm 0.21}$	/ 62789.10 \\
&DP = 10 (5)	&81.25	$_{\pm 0.47}$	/ 8054.98	&50.58	$_{\pm 0.46}$	/ 8055.43	&38.11	$_{\pm 0.22}$	/ 64441.45 \\
&DR = 0.2 (10)	&81.46	$_{\pm 0.29}$	/ 7848.44	&51.51	$_{\pm 0.32}$	/ 7848.88	&38.24	$_{\pm 0.39}$	/ 62789.10 \\
&DR = 0.5 (25)	&80.58	$_{\pm 0.41}$	/ 7228.83	&52.36	$_{\pm 0.36}$	/ 7229.23	&38.21	$_{\pm 0.28}$	/ 57832.07 \\
&DR = 0.7 (35)	&78.13	$_{\pm 0.55}$	/ 6815.75	&51.21	$_{\pm 0.28}$	/ 6816.13	&35.34	$_{\pm 0.29}$	/ 54527.38 \\ \hline
\multirow{11}{*}{\bf Mlp-Mixer  (\cite{mlp-mixer})}&Baseline (0)	&68.48	$_{\pm 0.36}$	/ 303.84	&37.39	$_{\pm 0.36}$	/ 304.76	&33.55	$_{\pm 0.19}$	/ 2811.49 \\
&Pe= 5 (20)	&68.90	$_{\pm 0.15}$	/ 253.71	&39.08	$_{\pm 0.19}$	/ 254.48	&34.51	$_{\pm 0.32}$	/ 2347.60 \\
&Pe= 10 (10)	&69.03	$_{\pm 0.40}$	/ 278.01	&38.52	$_{\pm 0.40}$	/ 278.86	&35.68	$_{\pm 0.39}$	/ 2572.52 \\
&Pr= 0.2 (20)	&69.03	$_{\pm 0.13}$	/ 267.38	&39.89	$_{\pm 0.13}$	/ 268.19	&36.02	$_{\pm 0.05}$	/ 2474.12 \\
&Pr= 0.5 (50)	&69.32	$_{\pm 0.17}$	/ 205.09	&40.18	$_{\pm 0.20}$	/ 205.71	&34.60	$_{\pm 0.11}$	/ 1897.76 \\
&Pr= 0.7 (70)	&68.93	$_{\pm 0.15}$	/ 180.79	&21.21	$_{\pm 0.30}$	/ 181.33	&25.42	$_{\pm 0.42}$	/ 1672.84 \\
&DP = 5 (10)	&64.13	$_{\pm 0.18}$	/ 288.65	&35.26	$_{\pm 0.19}$	/ 289.52	&28.31	$_{\pm 0.40}$	/ 2670.92 \\
&DP = 10 (5)	&65.35	$_{\pm 0.49}$	/ 296.25	&35.58	$_{\pm 0.11}$	/ 297.14	&27.27	$_{\pm 0.26}$	/ 2741.21 \\
&DR = 0.2 (10)	&67.15	$_{\pm 0.31}$	/ 288.65	&38.35	$_{\pm 0.16}$	/ 289.52	&29.18	$_{\pm 0.37}$	/ 2670.92 \\
&DR = 0.5 (25)	&66.15	$_{\pm 0.28}$	/ 265.86	&31.33	$_{\pm 0.29}$	/ 266.67	&31.47	$_{\pm 0.38}$	/ 2460.06 \\
&DR = 0.7 (35)	&64.05	$_{\pm 0.35}$	/ 250.67	&30.30	$_{\pm 0.21}$	/ 251.43	&25.55	$_{\pm 0.37}$	/ 2319.48 \\ \hline
\end{tabular}
    
    \label{tab:hypo_class}
\end{table}

Table~\ref{tab:hypo_class} presents our quantitative results, where the second column corresponds to the different sampling strategies. Each cell corresponds to the performance of the model ({\it Acc (\%)})  accompanied by the total number of Floating Point Operations Per Second (FLOPS) incurred throughout the model training process denoted in terms of Tera FLOPS (TFLOPS).
Upon meticulous examination of the table, it becomes apparent that periodic sampling every 10 epochs yields model performance indistinguishable from the baseline, achieving a commendable $10\%$ energy savings. Furthermore, employing sampling probabilities of 0.2 and 0.5 results in performances akin to the baselines, while achieving savings of approximately $20\%$ and $50\%$, respectively. Importantly, the performance drop remains consistent as energy savings increase from $20\%$ to $70\%$. These observations underscore the efficacy and versatility of our proposed sampling strategy, particularly in its applicability from the early stages of training, demonstrating improved efficiency as the training process unfolds. Additional experiments are detailed in the supplementary material under Section~\ref{sec:sstandard}.

\paragraph{Object Detection:} For the object detection task, we employed real-time object detectors such as YOLOv7 (\cite{yolov7}) and RT-DETR (\cite{rt-detr}), with quantitative results summarized in Table ~\ref{tab:obj_result}. All models achieve nearly identical performance compared to the baseline, facilitating substantial energy savings over the course of an entire epoch. Specifically, periodic sampling at intervals of 5 and 10 epochs results in energy savings of approximately $10\%$ and $20\%$, respectively, while maintaining nearly identical performance levels to the baseline. Moreover, even with random sampling, the achieved performance closely aligns with that of the baseline, underscoring the robustness and effectiveness of our proposed sampling strategies in simultaneously optimizing model performance and reducing energy consumption.
\begin{table}[!ht]
    
\tiny
\caption{Performance comparison of standard object detection models under diverse samplings, including periodic (Pe), and probabilistic (Pr) strategies over PascalVOC-2012 dataset. \label{tab:obj_result}}
    \centering
    \begin{tabular}{cl|cccc}
    \hline
         {\bf Model}&{\bf Strategy (\% savings)}&\textbf{\it mAP$_{.5}$}&\textbf{\it mAP$_{.5:.75}$}  \\
    \hline

\multirow{6}{*}{\textbf{YOLOv7  (\cite{yolov7})}}&Baseline (0)&	76.9&	59.7\\
&Pe= 5 (20) &	77.1&	60.1\\
&Pe= 10 (10) &	77.5&	60.3\\
&Pr= 0.2 (20) &	76.7&	59.6\\
&Pr= 0.5 (50) &	75.1&	57.3\\
&Pr= 0.7 (70) &	73.2&	54.6\\ \hline
\multirow{6}{*}{\textbf{RT-DETR  (\cite{rt-detr})}}&Baseline (0)&	78.4&	62.7\\
&Pe= 5 (20) &	77.2&	61.8\\
&Pe= 10 (10) &	78.6&	62.9\\
&Pr= 0.2 (20) &	78.7&	61.9\\
&Pr= 0.5 (50) &	77.4&	60.5\\
&Pr= 0.7 (70) &	76.1&	61.6\\ \hline

    \end{tabular}
\end{table}

\paragraph{Image Segmentation:} For the image segmentation task, various standard networks including U-Net (\cite{unet}), DeepLabv3 (\cite{Deeplabv3}), and Seg-Former (\cite{segformer}) were evaluated, with results summarized in Table ~\ref{tab:seg_result}. Periodic sampling at intervals of 5 and 10 epochs demonstrates remarkable similarity in performance compared to the baseline models, coupled with significant savings in backpropagation energy consumption of $20\%$ and $10\%$, respectively. These results highlight the promising energy efficiency gains attainable through our proposed approach.
\begin{table}[!ht]
\tiny
\caption{ Performance comparision of standard image segmentation models under diverse samplings, including periodic (Pe), probablistic (Pr) stategies over CityScapes dataset.}
    \label{tab:seg_result}
    \centering
    \begin{tabular}{cl|c}
    \hline
         {\bf Model}&{\bf Strategy (\% savings)}&{\it mIOU}  \\
         \hline
         \multirow{6}{*}{\bf DeepLabv3  (\cite{Deeplabv3})} &{Baseline (0)}&42.6\\
         &Pe= 5 (20) & 40.2 \\
         &Pe= 10 (10) & 42.5 \\
         &Pr= 0.2 (20) & 42.8 \\
         &Pr= 0.5 (50) & 41.5 \\
         &Pr= 0.7 (70) & 36.7 \\
         \hline
         \multirow{6}{*}{\bf UNet  (\cite{unet})}&{Baseline (0)}&	47.9\\
	&Pe= 5 (20) &	45.5\\
	&Pe= 10 (10) &	48.8\\
	&Pr= 0.2 (20) &	41.9\\
	&Pr= 0.5 (50) &	43.9\\
	&Pr= 0.7 (70) &	39.3\\ \hline
\multirow{6}{*}{\bf SegFormer  (\cite{segformer})}&	Baseline (0)&	62.5\\
	&Pe= 5 (20) &	60.2\\
	&Pe= 10 (10) &	62.3\\
	&Pr= 0.2 (20) &	62.8\\
	&Pr= 0.5 (50) &	62.1\\
	&Pr= 0.7 (70) &	58.5\\ \hline
		
    \end{tabular}
\end{table}

\paragraph{Domain Adaptation (DA) and Domain Generalization (DG):} In our experimental exploration of Domain Adaptation (DA) and Domain Generalization (DG) methodologies, we incorporated strategies such as MDD  (\cite{MDD}) and MCC  (\cite{MCC}) for DA, and techniques like VREx  (\cite{VREx}) and GroupDRO  (\cite{GroupDRO}) for DG, aimed at enhancing model robustness against unseen domains and improving generalization capabilities. Our experimental findings, detailed in Tables ~\ref{tab:DA_office} and ~\ref{tab:DG-office}, yielded compelling insights. Notably, significant energy savings of $20\%$ and $10\%$ were observed under periodic sampling scenarios with intervals of 5 and 10, respectively. Importantly, our proposed methodology demonstrated performance comparable to established baselines in these scenarios, showcasing its resilience and adaptability in the face of domain shifts. These findings validate the effectiveness of our approach. Further experiments are documented in the supplementary material under the section~\ref{sec:sdadg}.
\begin{table}[!ht]
\tiny
\caption{Performance of source ($S$) to target ($T$) ($S \rightarrow T$) Domain Adaptation (DA) methods on the Office-31  (\cite{office31}) dataset}
    \label{tab:DA_office}
    \centering
    \begin{tabular}{ll|cccccc|c}
    \hline
        {\bf Method} & {\bf Strategy (\% Savings)} & {\bf A $\rightarrow$ D} & {\bf A $\rightarrow$ W} & {\bf D $\rightarrow$ A} & {\bf D $\rightarrow$ W} & {\bf W $\rightarrow$ A} & {\bf W $\rightarrow$ D} & {\bf Avg}
\\ \hline
        \multirow{6}{*}{\bf MDD  (\cite{MDD})}& Baseline (0)&	94.2&	95.5&	76.3&	98.6&	72&	99.8&89.4\\
&Pe=5 (20)&	93&	94.7&	76.3&	99.8&	73.5&	99.8&89.5\\
&Pe=10 (10)&	93.4&	94.3&	75.8&	98.6&	73.4&	99.8&89.2\\
&Pr=0.2 (20)&	95&	95.1&	76.6&	98.5&	72.4&	99.8&89.5\\
&Pr=0.5 (50)&	93.2&	94.2&	75.7&	98.9&	72.4&	99.8&89\\
&Pr=0.7 (70)&	95&	93.6&	76.6&	98.5&	72.2&	99.8&89.3\\ \hline

\multirow{6}{*}{\bf MCC  (\cite{MCC})}&Baseline (0)&	95.6&	94.1&	75.5&	98.4&	74.2&	99.8&89.6\\
&Pe=5 (20)&	95.2&	94.3&	76.2&	98.5&	74.4&	99.6&89.7\\
&Pe=10 (10)&	94.8&	94.3&	76.1&	98.6&	74.8&	99.6&89.7\\
&Pr=0.2 (20)&	93.8&	94.7&	75.8&	98.6&	75.3&	99.8&89.6\\
&Pr=0.5 (50)&	93.8&	94.7&	75.8&	95.6&	75.3&	99.8&89.2\\
&Pr=0.7 (70)&	96&	94.7&	75.4&	92.6&	74.7&	99.8&88.9 \\ \hline
         
    \end{tabular}
\end{table}
\begin{table}[!ht]
\tiny
\caption{Performance across various Domain Generalization (DG) methods on the Office-Home  (\cite{officehome}) dataset.}
    \label{tab:DG-office}
    \centering
    \begin{tabular}{cl|cccc|c}
    \hline
      {\bf Method}  &{\bf Strategy (\% savings)} &{\bf A}&{\bf C}&{\bf P}&{\bf R}&{\bf Avg}  \\ \hline
        \multirow{6}{*}{\bf VREx  (\cite{VREx})} &Baseline(-)& 66.9&	54.9&	78.2&	80.9& 70.2\\
        & Pe = 5 (20)&66&55&78&80&69.9\\
        &Pe = 10 (10)&66.5&54.3&77.9&81&69.9\\
        &Pr = 0.2 (20)&67&54&79&80.5&70.1\\
        &Pr = 0.5 (50)&65&53&78&79&68.8\\
        &Pr = 0.7 (70)&67&53.6&77.3&78.5&69.1\\
        \hline
        \multirow{6}{*}{\bf GroupDRO  (\cite{GroupDRO})} &Baseline(-)& 66.7&	55.2&	78.8&	79.9& 70.0\\
        & Pe = 5 (20)&65&54.9&78.6&79.2&69.4\\
        &Pe = 10 (10)&67&55.9&78.3&80&69.9\\
        &Pr = 0.2 (20)&67&54.9&78.2&78.9&70.3\\
        &Pr = 0.5 (50)&66&54&78.3&78.5&69.2\\
        &Pr = 0.7 (70)&65&53.5&76.5&77.5&68.1\\
        \hline       
    \end{tabular}
\end{table}

\paragraph{Stochastic Federated Learning:} The quantitative results of the proposed Stochastic {\tt FedAvg} and {\tt FedProx} algorithms are presented in Table~\ref{tab:accuracy_fl_50}. This table summarizes the performance of our stochastic federated learning (FL) setting, where 50 local epochs are completed before each global round. In the fourth column, a period of 5 indicates that every fifth round triggers a global update through sampling from the Gaussian distribution, meaning the server does not aggregate client weights for this update, thereby saving on a communication round. Our experiments varied both the total number of clients and the number of clients sampled for a global update round. Table~\ref{tab:accuracy_fl_50} demonstrates that the proposed hypothesis achieves energy savings through a reduction in global rounds. Key findings include a slight performance drop with $20\%$ savings and almost identical performance to the baseline with $10\%$ savings across 50 global communication rounds. Additional experiments are reported in the supplementary material under the section~\ref{sec:sFL}.

In summary, our hypothesis that gradient elements can be modeled using a Gaussian distribution has been experimentally validated using several DL models spanning several computer vision tasks as well as in a distributed learning FL setting. Importantly, the experiments demonstrate that DL models can be trained at a lower energy footprint without loss in performance.  
\begin{table}[!ht]
\tiny
    \caption{Performance evaluation of the Stochastic FL setting over 20 global rounds, accounting for Total Clients (TC) and Selected Clients (SC) per round, with periodic (Pe) sampling, trained over 50 local epochs.}
    \label{tab:accuracy_fl_50}
    \centering
    \begin{tabular}{c|ccl|cc|cc|cc|cc}
        \hline
        \multirow{2}{*}{\bf Method}&\multirow{2}{*}{\bf TC}& \multirow{2}{*}{\bf SC}&\multirow{2}{*}{\bf Strategy (\%savings)}& \multicolumn{2}{c}{\bf MobileNetV2}& \multicolumn{2}{c}{\bf ShuffleNetV2}& \multicolumn{2}{c}{\bf SqueezeNet}& \multicolumn{2}{c}{\bf ResNet-50} \\ 
        &&&&{\bf CIFAR-10}&{\bf CIFAR-100}&{\bf CIFAR-10}&{\bf CIFAR-100}&{\bf CIFAR-10}&{\bf CIFAR-100}&{\bf CIFAR-10}&{\bf CIFAR-100}\\
        \hline
        \multirow{12}{*}{\bf {\tt FedAvg}}&\multirow{3}{*}{\bf 5}&\multirow{3}{*}{\bf 1}&Baseline (0)&70.04&	30.69&	57.41&	26.72&	75.58&	28.42&	75.63&	36.18\\
&&&Pe= 5 (20)&67.14&	32.23&	56.68&	24.85&	75.48&	26.09	&70.23	&35.12\\
&&&Pe= 10 (10)&71.13&	27.2&	54.95&	26.09&	76.33&	27.07&	73.24&	36.11\\ \cline{2-12}
&\multirow{3}{*}{\bf 5}&\multirow{3}{*}{\bf 2}&Baseline (0)&76.13&	38.18&	69.25&	33.01&	76.13&	35.28&	77.53&	40.96\\
&&&Pe= 5 (20)&74.65&	31.06&	65.43&	31.07&	74.65&	33.07&	75.16&	37.57\\
&&&Pe= 10 (10)&74.91&	35&	63.74&	29.91&	74.91&	32.66&	77.27&	40.97\\ \cline{2-12}
&\multirow{3}{*}{\bf 10}&\multirow{3}{*}{\bf 2}&Baseline (0)&67.91&	27.2&	57.98&	23.55&	76.22&	27.41&	69.31&	32.6\\
&&&Pe= 5 (20)&64.91&	24.23&	57.03&	21.52&	72.47&	26.07&	66.24&	29.43\\
&&&Pe= 10 (10)&69.18&	26.81&	60.3&	24.03&	75.51&	24.99&	66.9&	31.43\\ \cline{2-12}
&\multirow{3}{*}{\bf 10}&\multirow{3}{*}{\bf 5}&Baseline (0)&71.41&	27.66&	63.06&	28.95&	77.93&	31.37&	69.55&	35.98 \\
&&&Pe= 5 (20)&68.54&	26.81&	62.35&	28.07&	76.03&	29.23&	68.84&	31.85\\
&&&Pe= 10 (10)&70.56&	29.33&	62.99&	29.47&	78.28&	30.15&	70.03&	33.08\\ \hline
\multirow{12}{*}{\bf {\tt FedProx}}&\multirow{3}{*}{\bf 5}&\multirow{3}{*}{\bf 1}& Baseline (0)& 71.41&	32.52&	58.2&	27.15&	76.32&	28.68&	75.91&	37.12\\
&&&Pe= 5 (20)&68.25&	30.51&	57.9&	25.61&	76.25&	26.91	&70.08&	36.11\\
&&&Pe= 10 (10)&72.1&	29.91&	56.85&	27.02&	78.45&	27.99&	73.85&	37.06\\ \cline{2-12}
&\multirow{3}{*}{\bf 5}&\multirow{3}{*}{\bf 2}&Baseline (0)&76.15&	39.25&	70.08&	34.01&	78.05&	36.23&	78.65&	41.05\\
&&&Pe= 5 (20)&74.99&	32.15&	68.25&	33.09&	75.65&	34.23&	75.91&	38.05\\
&&&Pe= 10 (10)&75.05&	33.05&	67.45&	32.51&	75.99&	33.07&	77.74&	41.01 \\ \cline{2-12}
&\multirow{3}{*}{\bf 10}&\multirow{3}{*}{\bf 2}&Baseline (0)&68.91&	29.16&	60.25&	24.14&	76.91&	27.66&	70.55&	33.47\\
&&&Pe= 5 (20)&66.12&	27.45&	60.02&	22.05&	74.65&	26.81&	69.84&	31.35\\
&&&Pe= 10 (10)&69.05&	28.07&	61.25&	25.01&	77.67&	25.43&	69.03&	32.65\\ \cline{2-12}
&\multirow{3}{*}{\bf 10}&\multirow{3}{*}{\bf 5}&Baseline (0)&72.05&	29.92&	63.55&	28.99&	77.99&	32.23&	70.95&	38.18\\
&&&Pe= 5 (20)&69.15&	28.15&	63.01&	27.45&	76.92&	30.65&	69.14&	32.74\\
&&&Pe= 10 (10)&71.25&	30.95&	63.75&	30.09&	78.99&	29.16&	70.99&	36.65\\ \hline

    \end{tabular}
\end{table}
\paragraph{Limitations:} A limitation of this work is the simplifying assumption about the parameter errors in a given layer of a DL model being i.i.d. While we found this assumption to hold in our experiments, we plan to work on a more rigorous justification for it.

\section{Conclusion}
We introduced a simple yet effective gradient sampling technique {\em GradSamp} to reduce energy consumption in practical Deep Learning (DL) models across computer vision tasks like image classification, object detection, and image segmentation. We evaluated its performance in challenging scenarios such as Domain Adaptation (DA) and Domain Generalization (DG), showing its effectiveness even in out-of-distribution situations. Additionally, we demonstrated its applicability in Federated Learning (FL) settings, emphasizing its practicality in decentralized environments. Comparative analyses across various training setups confirmed the robustness and practicality of our approach. Overall, our findings indicate that even a basic periodic sampling strategy can achieve significant energy savings without compromising model performance.
\bibliography{arxiv}

\begin{thebibliography}{45}
\providecommand{\natexlab}[1]{#1}
\providecommand{\url}[1]{\texttt{#1}}
\expandafter\ifx\csname urlstyle\endcsname\relax
  \providecommand{\doi}[1]{doi: #1}\else
  \providecommand{\doi}{doi: \begingroup \urlstyle{rm}\Url}\fi

\bibitem[KUO(2023)]{KUO2023103685}
Green learning: Introduction, examples and outlook.
\newblock \emph{Journal of Visual Communication and Image Representation},
  90:\penalty0 103685, 2023.
\newblock ISSN 1047-3203.
\newblock \doi{https://doi.org/10.1016/j.jvcir.2022.103685}.

\bibitem[Chen et~al.(2017)Chen, Papandreou, Schroff, and Adam]{Deeplabv3}
Liang-Chieh Chen, George Papandreou, Florian Schroff, and Hartwig Adam.
\newblock Rethinking atrous convolution for semantic image segmentation.
\newblock \emph{arXiv preprint arXiv:1706.05587}, 2017.

\bibitem[Chen \& Shao(2004)Chen and Shao]{chen2004normal}
Louis H.~Y. Chen and Qi-Man Shao.
\newblock Normal approximation under local dependence.
\newblock \emph{The Annals of Probability}, 32\penalty0 (3):\penalty0
  1985--2028, 2004.
\newblock ISSN 00911798.
\newblock URL \url{http://www.jstor.org/stable/3481601}.

\bibitem[Cordts et~al.(2016)Cordts, Omran, Ramos, Rehfeld, Enzweiler, Benenson,
  Franke, Roth, and Schiele]{cityscapes}
Marius Cordts, Mohamed Omran, Sebastian Ramos, Timo Rehfeld, Markus Enzweiler,
  Rodrigo Benenson, Uwe Franke, Stefan Roth, and Bernt Schiele.
\newblock The cityscapes dataset for semantic urban scene understanding.
\newblock In \emph{Proceedings of the IEEE conference on computer vision and
  pattern recognition}, pp.\  3213--3223, 2016.

\bibitem[D'agostino \& Pearson(1973)D'agostino and Pearson]{pearson}
RALPH D'agostino and Egon~S Pearson.
\newblock Tests for departure from normality. empirical results for the
  distributions of $b_2$ and $\sqrt{b_1}$.
\newblock \emph{Biometrika}, 60\penalty0 (3):\penalty0 613--622, 1973.

\bibitem[Desislavov et~al.(2023)Desislavov, Plumed, and
  Hernandez-Orallo]{energy}
Radosvet Desislavov, Fernando Plumed, and Jose Hernandez-Orallo.
\newblock Trends in ai inference energy consumption: Beyond the
  performance-vs-parameter laws of deep learning.
\newblock \emph{Sustainable Computing: Informatics and Systems}, 38:\penalty0
  100857, 02 2023.
\newblock \doi{10.1016/j.suscom.2023.100857}.

\bibitem[Esser et~al.(2015)Esser, Appuswamy, Merolla, Arthur, and
  Modha]{esser2015backpropagation}
Steve~K Esser, Rathinakumar Appuswamy, Paul Merolla, John~V Arthur, and
  Dharmendra~S Modha.
\newblock Backpropagation for energy-efficient neuromorphic computing.
\newblock \emph{Advances in neural information processing systems}, 28, 2015.

\bibitem[Everingham et~al.()Everingham, Van~Gool, Williams, Winn, and
  Zisserman]{pascal12}
M.~Everingham, L.~Van~Gool, C.~K.~I. Williams, J.~Winn, and A.~Zisserman.
\newblock The {PASCAL} {V}isual {O}bject {C}lasses {C}hallenge 2012 {(VOC2012)}
  {R}esults.
\newblock
  http://www.pascal-network.org/challenges/VOC/voc2012/workshop/index.html.

\bibitem[Francazi et~al.(2023)Francazi, Baity-Jesi, and
  Lucchi]{pmlr-v202-francazi23a}
Emanuele Francazi, Marco Baity-Jesi, and Aurelien Lucchi.
\newblock A theoretical analysis of the learning dynamics under class
  imbalance.
\newblock In Andreas Krause, Emma Brunskill, Kyunghyun Cho, Barbara Engelhardt,
  Sivan Sabato, and Jonathan Scarlett (eds.), \emph{Proceedings of the 40th
  International Conference on Machine Learning}, volume 202 of
  \emph{Proceedings of Machine Learning Research}, pp.\  10285--10322. PMLR,
  23--29 Jul 2023.

\bibitem[He et~al.(2016)He, Zhang, Ren, and Sun]{Resnet}
Kaiming He, Xiangyu Zhang, Shaoqing Ren, and Jian Sun.
\newblock Deep residual learning for image recognition.
\newblock In \emph{Proceedings of the IEEE conference on computer vision and
  pattern recognition}, pp.\  770--778, 2016.

\bibitem[Hoefler et~al.(2021)Hoefler, Alistarh, Ben-Nun, Dryden, and
  Peste]{hoefler2021sparsity}
Torsten Hoefler, Dan Alistarh, Tal Ben-Nun, Nikoli Dryden, and Alexandra Peste.
\newblock Sparsity in deep learning: Pruning and growth for efficient inference
  and training in neural networks.
\newblock \emph{The Journal of Machine Learning Research}, 22\penalty0
  (1):\penalty0 10882--11005, 2021.

\bibitem[Iandola et~al.(2016{\natexlab{a}})Iandola, Han, Moskewicz, Ashraf,
  Dally, and Keutzer]{iandola2016squeezenet}
Forrest~N Iandola, Song Han, Matthew~W Moskewicz, Khalid Ashraf, William~J
  Dally, and Kurt Keutzer.
\newblock Squeezenet: Alexnet-level accuracy with 50x fewer parameters and< 0.5
  mb model size.
\newblock \emph{arXiv preprint arXiv:1602.07360}, 2016{\natexlab{a}}.

\bibitem[Iandola et~al.(2016{\natexlab{b}})Iandola, Han, Moskewicz, Ashraf,
  Dally, and Keutzer]{squeezenet}
Forrest~N Iandola, Song Han, Matthew~W Moskewicz, Khalid Ashraf, William~J
  Dally, and Kurt Keutzer.
\newblock Squeezenet: Alexnet-level accuracy with 50x fewer parameters and< 0.5
  mb model size.
\newblock \emph{arXiv preprint arXiv:1602.07360}, 2016{\natexlab{b}}.

\bibitem[Jin et~al.(2020)Jin, Wang, Long, and Wang]{MCC}
Ying Jin, Ximei Wang, Mingsheng Long, and Jianmin Wang.
\newblock Minimum class confusion for versatile domain adaptation.
\newblock In \emph{Computer Vision--ECCV 2020: 16th European Conference,
  Glasgow, UK, August 23--28, 2020, Proceedings, Part XXI 16}, pp.\  464--480.
  Springer, 2020.

\bibitem[Krizhevsky et~al.(2009)Krizhevsky, Hinton, et~al.]{cifar10}
Alex Krizhevsky, Geoffrey Hinton, et~al.
\newblock Learning multiple layers of features from tiny images.
\newblock 2009.

\bibitem[Krueger et~al.(2021)Krueger, Caballero, Jacobsen, Zhang, Binas, Zhang,
  Priol, and Courville]{VREx}
David Krueger, Ethan Caballero, Joern-Henrik Jacobsen, Amy Zhang, Jonathan
  Binas, Dinghuai Zhang, Remi~Le Priol, and Aaron Courville.
\newblock Out-of-distribution generalization via risk extrapolation (rex).
\newblock In \emph{ICML}, 2021.

\bibitem[Li et~al.(2017)Li, Yang, Song, and Hospedales]{pacs}
Da~Li, Yongxin Yang, Yi-Zhe Song, and Timothy~M Hospedales.
\newblock Deeper, broader and artier domain generalization.
\newblock In \emph{Proceedings of the IEEE international conference on computer
  vision}, pp.\  5542--5550, 2017.

\bibitem[Li et~al.(2020{\natexlab{a}})Li, Sahu, Talwalkar, and
  Smith]{fed_intro}
Tian Li, Anit~Kumar Sahu, Ameet Talwalkar, and Virginia Smith.
\newblock Federated learning: Challenges, methods, and future directions.
\newblock \emph{IEEE signal processing magazine}, 37\penalty0 (3):\penalty0
  50--60, 2020{\natexlab{a}}.

\bibitem[Li et~al.(2020{\natexlab{b}})Li, Sahu, Zaheer, Sanjabi, Talwalkar, and
  Smith]{fedprox}
Tian Li, Anit~Kumar Sahu, Manzil Zaheer, Maziar Sanjabi, Ameet Talwalkar, and
  Virginia Smith.
\newblock Federated optimization in heterogeneous networks.
\newblock \emph{Proceedings of Machine learning and systems}, 2:\penalty0
  429--450, 2020{\natexlab{b}}.

\bibitem[Lin et~al.(2022)Lin, Zhou, You, Rao, and Kuo]{lin2022geometrical}
Ruiyuan Lin, Zhiruo Zhou, Suya You, Raghuveer Rao, and C-C~Jay Kuo.
\newblock Geometrical interpretation and design of multilayer perceptrons.
\newblock \emph{IEEE Transactions on Neural Networks and Learning Systems},
  2022.

\bibitem[Liu et~al.(2022)Liu, Zhu, and Belkin]{liu2022loss}
Chaoyue Liu, Libin Zhu, and Mikhail Belkin.
\newblock Loss landscapes and optimization in over-parameterized non-linear
  systems and neural networks.
\newblock \emph{Applied and Computational Harmonic Analysis}, 59:\penalty0
  85--116, 2022.

\bibitem[Liu et~al.(2023)Liu, Han, Ma, Zhang, Yang, Tian, He, Li, He, Liu,
  et~al.]{liu2023summary}
Yiheng Liu, Tianle Han, Siyuan Ma, Jiayue Zhang, Yuanyuan Yang, Jiaming Tian,
  Hao He, Antong Li, Mengshen He, Zhengliang Liu, et~al.
\newblock Summary of chatgpt/gpt-4 research and perspective towards the future
  of large language models.
\newblock \emph{arXiv preprint arXiv:2304.01852}, 2023.

\bibitem[Liu et~al.(2021)Liu, Lin, Cao, Hu, Wei, Zhang, Lin, and Guo]{swin}
Ze~Liu, Yutong Lin, Yue Cao, Han Hu, Yixuan Wei, Zheng Zhang, Stephen Lin, and
  Baining Guo.
\newblock Swin transformer: Hierarchical vision transformer using shifted
  windows.
\newblock In \emph{Proceedings of the IEEE/CVF international conference on
  computer vision}, pp.\  10012--10022, 2021.

\bibitem[Loshchilov \& Hutter(2017)Loshchilov and
  Hutter]{loshchilov2017decoupled}
Ilya Loshchilov and Frank Hutter.
\newblock Decoupled weight decay regularization.
\newblock \emph{arXiv preprint arXiv:1711.05101}, 2017.

\bibitem[Ma et~al.(2018{\natexlab{a}})Ma, Zhang, Zheng, and Sun]{Ma_2018_ECCV}
Ningning Ma, Xiangyu Zhang, Hai-Tao Zheng, and Jian Sun.
\newblock Shufflenet v2: Practical guidelines for efficient cnn architecture
  design.
\newblock In \emph{Proceedings of the European Conference on Computer Vision
  (ECCV)}, September 2018{\natexlab{a}}.

\bibitem[Ma et~al.(2018{\natexlab{b}})Ma, Zhang, Zheng, and Sun]{shufflenetv2}
Ningning Ma, Xiangyu Zhang, Hai-Tao Zheng, and Jian Sun.
\newblock Shufflenet v2: Practical guidelines for efficient cnn architecture
  design.
\newblock In \emph{Proceedings of the European conference on computer vision
  (ECCV)}, pp.\  116--131, 2018{\natexlab{b}}.

\bibitem[Malladi et~al.(2023)Malladi, Gao, Nichani, Damian, Lee, Chen, and
  Arora]{mezo}
Sadhika Malladi, Tianyu Gao, Eshaan Nichani, Alex Damian, Jason~D Lee, Danqi
  Chen, and Sanjeev Arora.
\newblock Fine-tuning language models with just forward passes.
\newblock \emph{arXiv preprint arXiv:2305.17333}, 2023.

\bibitem[McMahan et~al.(2017)McMahan, Moore, Ramage, Hampson, and
  y~Arcas]{fedavg}
Brendan McMahan, Eider Moore, Daniel Ramage, Seth Hampson, and Blaise~Aguera
  y~Arcas.
\newblock Communication-efficient learning of deep networks from decentralized
  data.
\newblock In \emph{Artificial intelligence and statistics}, pp.\  1273--1282.
  PMLR, 2017.

\bibitem[Merolla et~al.(2014)Merolla, Arthur, Alvarez-Icaza, Cassidy, Sawada,
  Akopyan, Jackson, Imam, Guo, Nakamura, et~al.]{merolla2014million}
Paul~A Merolla, John~V Arthur, Rodrigo Alvarez-Icaza, Andrew~S Cassidy, Jun
  Sawada, Filipp Akopyan, Bryan~L Jackson, Nabil Imam, Chen Guo, Yutaka
  Nakamura, et~al.
\newblock A million spiking-neuron integrated circuit with a scalable
  communication network and interface.
\newblock \emph{Science}, 345\penalty0 (6197):\penalty0 668--673, 2014.

\bibitem[Peng et~al.(2017)Peng, Usman, Kaushik, Hoffman, Wang, and
  Saenko]{visda}
Xingchao Peng, Ben Usman, Neela Kaushik, Judy Hoffman, Dequan Wang, and Kate
  Saenko.
\newblock Visda: The visual domain adaptation challenge.
\newblock \emph{ArXiv}, abs/1710.06924, 2017.
\newblock URL \url{https://api.semanticscholar.org/CorpusID:28698351}.

\bibitem[Ronneberger et~al.(2015)Ronneberger, Fischer, and Brox]{unet}
Olaf Ronneberger, Philipp Fischer, and Thomas Brox.
\newblock U-net: Convolutional networks for biomedical image segmentation.
\newblock In \emph{Medical Image Computing and Computer-Assisted
  Intervention--MICCAI 2015: 18th International Conference, Munich, Germany,
  October 5-9, 2015, Proceedings, Part III 18}, pp.\  234--241. Springer, 2015.

\bibitem[Rumelhart et~al.(1986)Rumelhart, Hinton, and Williams]{backprop}
David~E Rumelhart, Geoffrey~E Hinton, and Ronald~J Williams.
\newblock Learning representations by back-propagating errors.
\newblock \emph{nature}, 323\penalty0 (6088):\penalty0 533--536, 1986.

\bibitem[Saenko et~al.(2010)Saenko, Kulis, Fritz, and Darrell]{office31}
Kate Saenko, Brian Kulis, Mario Fritz, and Trevor Darrell.
\newblock Adapting visual category models to new domains.
\newblock In \emph{European Conference on Computer Vision}, 2010.
\newblock URL \url{https://api.semanticscholar.org/CorpusID:7534823}.

\bibitem[Sagawa et~al.(2020)Sagawa, Koh, Hashimoto, and Liang]{GroupDRO}
Shiori Sagawa, Pang~Wei Koh, Tatsunori~B. Hashimoto, and Percy Liang.
\newblock Distributionally robust neural networks for group shifts: On the
  importance of regularization for worst-case generalization.
\newblock In \emph{ICLR}, 2020.

\bibitem[Sandler et~al.(2018)Sandler, Howard, Zhu, Zhmoginov, and
  Chen]{mobilenetv2}
Mark Sandler, Andrew Howard, Menglong Zhu, Andrey Zhmoginov, and Liang-Chieh
  Chen.
\newblock Mobilenetv2: Inverted residuals and linear bottlenecks.
\newblock In \emph{Proceedings of the IEEE conference on computer vision and
  pattern recognition}, pp.\  4510--4520, 2018.

\bibitem[Simonyan \& Zisserman(2014)Simonyan and Zisserman]{simonyan2014very}
Karen Simonyan and Andrew Zisserman.
\newblock Very deep convolutional networks for large-scale image recognition.
\newblock \emph{arXiv preprint arXiv:1409.1556}, 2014.

\bibitem[Strubell et~al.(2020)Strubell, Ganesh, and
  McCallum]{Strubell_Ganesh_McCallum_2020}
Emma Strubell, Ananya Ganesh, and Andrew McCallum.
\newblock Energy and policy considerations for modern deep learning research.
\newblock \emph{Proceedings of the AAAI Conference on Artificial Intelligence},
  34\penalty0 (09):\penalty0 13693--13696, Apr. 2020.
\newblock \doi{10.1609/aaai.v34i09.7123}.

\bibitem[Sze et~al.(2017)Sze, hsin Chen, Yang, and Emer]{Sze2017EfficientPO}
Vivienne Sze, Yu~hsin Chen, Tien-Ju Yang, and Joel~S. Emer.
\newblock Efficient processing of deep neural networks: A tutorial and survey.
\newblock \emph{Proceedings of the IEEE}, 105:\penalty0 2295--2329, 2017.

\bibitem[Tavanaei(2020)]{tiny}
Amirhossein Tavanaei.
\newblock Embedded encoder-decoder in convolutional networks towards
  explainable ai.
\newblock \emph{arXiv preprint arXiv:2007.06712}, 2020.

\bibitem[Tolstikhin et~al.(2021)Tolstikhin, Houlsby, Kolesnikov, Beyer, Zhai,
  Unterthiner, Yung, Steiner, Keysers, Uszkoreit, et~al.]{mlp-mixer}
Ilya~O Tolstikhin, Neil Houlsby, Alexander Kolesnikov, Lucas Beyer, Xiaohua
  Zhai, Thomas Unterthiner, Jessica Yung, Andreas Steiner, Daniel Keysers,
  Jakob Uszkoreit, et~al.
\newblock Mlp-mixer: An all-mlp architecture for vision.
\newblock \emph{Advances in neural information processing systems},
  34:\penalty0 24261--24272, 2021.

\bibitem[Venkateswara et~al.(2017)Venkateswara, Eusebio, Chakraborty, and
  Panchanathan]{officehome}
Hemanth Venkateswara, Jose Eusebio, Shayok Chakraborty, and Sethuraman
  Panchanathan.
\newblock Deep hashing network for unsupervised domain adaptation.
\newblock In \emph{IEEE Conf. on Computer Vision and Pattern Recognition
  ({CVPR})}, 2017.

\bibitem[Wang et~al.(2023)Wang, Bochkovskiy, and Liao]{yolov7}
Chien-Yao Wang, Alexey Bochkovskiy, and Hong-Yuan~Mark Liao.
\newblock Yolov7: Trainable bag-of-freebies sets new state-of-the-art for
  real-time object detectors.
\newblock In \emph{Proceedings of the IEEE/CVF Conference on Computer Vision
  and Pattern Recognition}, pp.\  7464--7475, 2023.

\bibitem[Xie et~al.(2021)Xie, Wang, Yu, Anandkumar, Alvarez, and
  Luo]{segformer}
Enze Xie, Wenhai Wang, Zhiding Yu, Anima Anandkumar, Jose~M Alvarez, and Ping
  Luo.
\newblock Segformer: Simple and efficient design for semantic segmentation with
  transformers.
\newblock \emph{Advances in Neural Information Processing Systems},
  34:\penalty0 12077--12090, 2021.

\bibitem[Zhang et~al.(2019)Zhang, Liu, Long, and Jordan]{MDD}
Yuchen Zhang, Tianle Liu, Mingsheng Long, and Michael Jordan.
\newblock Bridging theory and algorithm for domain adaptation.
\newblock In \emph{ICML}, 2019.

\bibitem[Zhao et~al.(2023)Zhao, Lv, Xu, Wei, Wang, Dang, Liu, and
  Chen]{rt-detr}
Yian Zhao, Wenyu Lv, Shangliang Xu, Jinman Wei, Guanzhong Wang, Qingqing Dang,
  Yi~Liu, and Jie Chen.
\newblock Detrs beat yolos on real-time object detection, 2023.

\end{thebibliography}
\bibliographystyle{iclr2024_conference}
\newpage
\appendix
\section{Experimental Details} \label{sec:sup_expdetails}
We implement the different modelling tasks by incorporating the hyperparameters as follows. For the image classification task, we experimented over 200 epochs with a batch size of 32 for CIFAR-10, CIFAR-100 datasets (resized to $32\times 32$) and batch size of 100 for Tiny ImageNet dataset (resized to $64\times 64$). We use SGD for gradient estimation, a learning rate of 0.001. and a momentum of 0.9 that facilitates convergence by integrating the past update directions. Finally, we use a weight decay of 0.001 to prevent overfitting. Additionally, we included a dropout layer for MobileNetV2  (\cite{mobilenetv2}), ShuffleNetV2  (\cite{shufflenetv2}) with a dropout rate of 0.2, and SqueezeNet  (\cite{squeezenet}) with a dropout rate of 0.5 which encourages model robustness by reducing the models’ reliance on specific features. For the Swin Transformer, we use a patch size of 2, a window size of 4, an attention head patch probability of 0.3, and 32 attention heads. We use the cross-entropy loss to assess the loss performance of the models. For the object detection task, we trained for 100 epochs with a batch size of 16 (resized to $416\times  416$) for both yolov7  (\cite{yolov7}) and Rt-Detr  (\cite{rt-detr}). For the image segmentation task, we use ResNet-50  (\cite{Resnet}) as the backbone for training UNet  (\cite{unet}) and DeeplabV3  (\cite{Deeplabv3}). The training was done for 100 epochs with a  batch of size 32 (resized to 256x512). We use AdamW  (\cite{loshchilov2017decoupled}) optimizer with a learning rate of 0.001. Finally, for Domain Adaptation (DA) and Domain Generalization (DG), we utilized ResNet-50 as the foundational architecture, known for its robustness in various computer vision tasks. Our training regimen consisted of 50 epochs, with each epoch comprising 1000 and 500 iterations for DA and DG, respectively, ensuring a comprehensive exploration of the model's learning space. The experiments are performed using an ``NVIDIA Tesla V100-SXM2-32GB`` GPU. 

\section{Standard Experiments}\label{sec:sstandard}

\begin{table*}[!ht]
    \centering
    \tiny
    \caption{Performance comparision of standard models under diverse sampling, including periodic (Pe), probabilistic (Pr), Delayed Period
(DP), Delayed Random (DR) samplings along with total FLOPS (in Tera FLOPS).}
    \begin{tabular}{ll|ccc}
    \hline
         {\bf Model}& {\bf Strategy (\% savings)} & {\bf CIFAR-10/ TFLOPS} & {\bf CIFAR-100/ TFLOPS} & {\bf TINY/ TFLOPS} \\ \hline
         \multirow{11}{*}{\bf MobileNetV2  (\cite{mobilenetv2})}& Baseline (0)&81.58/ 273.00&39.24/ 273.46
&33.15/ 2146.29
 \\
         & Pe= 5 (20) & 80.78/ 227.96 &38.18/ 228.34
&32.98/ 1792.15
 \\
         & Pe= 10 (10) & 79.96/ 249.80&38.47/ 250.22
&33.67/ 1963.86
 \\
         & Pr= 0.2 (20)& 80.66/ 240.24&39.01/ 240.65
&33.7/ 1888.74
 \\
         & Pr= 0.5 (50)& 79.53/ 184.28&40.9/ 184.59
&33.4/ 1448.75
 \\
         & Pr= 0.7 (70)& 73.54/ 162.44&34/ 162.71
&29.17/ 1277.04
 \\
&DP= 5 (10)&77.77/ 251.16	&38.04/ 251.59	&33.51/ 1973.96 \\
&DP= 10 (5)&78.7/ 260.72&	36.46/ 261.16	&32.83/ 2049.06 \\
&DR= 0.2 (10)&77.53/ 247.07&	35.97/ 247.49	&33.20/ 1941.78 \\
&DR= 0.5 (25)&76.24/ 229.32&	37.31/ 229.71	&33.85/ 1802.31 \\
&DR= 0.7 (35)&76.94/ 218.40&	37.94/ 218.77	&34.58/ 1716.49\\ \hline
\multirow{11}{*}{\bf ShuffleNetV2  (\cite{Ma_2018_ECCV})} & Baseline (0)&69.87/ 129.18&34.77/  129.64
&36.48/ 1003.55
 \\
&Pe= 5 (20)&73.58/ 107.86&34.4/ 108.25
&36.02/ 837.96
\\
&Pe= 10 (10)&73.66/ 118.19&35/ 118.62
&37.26/ 918.24
\\
&Pr= 0.2 (20)&72.49/ 113.67&35.23/ 114.08
&35.78/ 883.12
\\
&Pr= 0.5 (50)&74.03/ 87.19&36.66/ 87.50
&30.45/ 677.39
\\
&Pr= 0.7 (70)&69.8/ 76.86&32.49/ 77.13
&26.47/ 597.11
\\
&DP= 5 (10)&70.54/ 118.84&	35.11/ 6453.10&	31.69/ 922.97\\
&DP= 10 (5)&69.82/ 123.36&	34.3/ 123.80	&30.88/ 958.09\\
&DR= 0.2 (10)&70.72/ 116.90&	34.67/ 117.32&	31.15/ 907.92\\
&DR= 0.5 (25)&71.63/ 108.51	&34.50/ 108.89&	33.54/ 842.71\\
&DR= 0.7 (35)&70.45/ 103.34	&35.54/ 103.71&	33.24/ 802.58\\
\hline
\multirow{11}{*}{\bf SqueezeNet  (\cite{iandola2016squeezenet})} & Baseline (0)&69.87/ 129.18&34.77/  129.64
&36.48/ 1003.55
 \\
&Pe= 5 (20)&73.58/ 107.86&34.4/ 108.25
&36.02/ 837.96
 \\
&Pe= 10 (10)&73.66/ 118.19&35/ 118.62
&37.26/ 918.24
\\
&Pr= 0.2 (20)&72.49/ 113.67&35.23/ 114.08
&35.78/ 883.12
\\
&Pr= 0.5 (50)&74.03/ 87.19&36.66/ 87.50
&30.45/ 677.39
\\
&Pr= 0.7 (70)&69.8/ 76.86&32.49/ 77.13
&26.47/ 597.11
\\
&DP= 5 (10)&77.73/ 336.63&	37.2/ 337.15&	37.67/ 3635.84 \\
&DP= 10 (5)&78.36/ 349.44&	39/ 349.98&	37.59/ 3774.16 \\
&DR= 0.2 (10)&77.43/ 331.15&	39.62/ 331.65&	38.50/ 3576.56 \\
&DR= 0.5 (25)&76.44/ 307.36&	38.2/ 307.83&	33.7/ 3319.68 \\
&DR= 0.7 (35)&77.07/ 292.73&	40.54/ 293.17&	35.57/ 3161.60 \\
\hline
\multirow{11}{*}{\bf VGG-16  (\cite{simonyan2014very})}&Baseline (0)&88.7/ 4329.39&	56.85/ 4333.08&	34.78/ 27472.19 \\
&Pe= 5 (20)&87.69/ 3615.04&	54.92/ 3618.12&	35.76/ 22939.28\\
&Pe= 10 (10)&87.8/ 3961.39&	52.05/ 3964.76&	35.35/ 25137.05\\
&Pr= 0.2 (20)&88.45/ 3809.86&	51.6/ 3813.11&	35.25/ 24175.52\\
&Pr= 0.5 (50)&87.26/ 2922.34&	52.44/ 2924.82&	34.27/ 18543.72\\
&Pr= 0.7 (70)&85.77/ 2575.98&	51.69/  2578.18&	36.9/ 16345.95\\
&DP= 5 (10)&82.90/ 3983.04&	42.71/ 3986.43&	35.22/ 25266.33\\
&DP= 10 (5)&84.12/ 4134.56&	48.43/ 4138.09&	35/ 26227.55\\
&DR= 0.2 (10)&83.03/ 3918.10&	44.22/ 3921.43&	35.69/ 24854.38\\
&DR= 0.5 (25)&84.76/ 3636.68&	43.63/ 3639.78&	36.05/ 23069.26\\
&DR= 0.7 (35)&84.36/ 3463.51&	43.31/ 3466.46&	36.31/ 21970.72\\
\hline
    \end{tabular} 
    
    \label{tab:sNon-FL}
\end{table*}

\section{Comparitive Analysis}
\subsection{Actual vs Effective backprops}
An important comparison involved training the baseline model using the same number of actual backpropagation operations as our method, but without sampling. These results are presented in Table~\ref{tab:effective} and ~\ref{tab:snodecay}. All these experimental results further demonstrate the efficacy of our algorithm.
\begin{table}[!htb]
    \centering
    \caption{Performance of models trained with the same number of {\em actual backprop} operations as the sampling method uses.}
    \label{tab:effective}
    \tiny
    \begin{tabular}{lc|cc| cc| cc}
    \hline
         \multirow{2}{*}{\bf Model}& \multirow{2}{*}{\bf Actual \# Backprops} & \multicolumn{2}{c|}{\bf CIFAR-10} & \multicolumn{2}{c|}{\bf CIFAR-100} & \multicolumn{2}{c}{\bf TINY}\\ \cline{3-8}
         &&\textbf{Sampling} &\textbf{No Sampling}&\textbf{Sampling} &\textbf{No Sampling}&\textbf{Sampling} &\textbf{No Sampling} \\ \hline

\multirow{5}{*}{\bf ResNet-50  (\cite{Resnet})}&167 &80.13& 79.63&	38.37& 36.76&	29.28& 28.17 \\
&183 &79.43& 77.61&	40.17& 38.67&	33.2& 33.26\\
&176 &80.16& 78.96&	40.76& 40.28&	30.37& 28.56\\
&135&79.57& 77.26&	39.03& 38.96&	27.78& 25.06\\
&119&75.93& 74.96&	40.82& 39.97&	29.21& 27.58\\

\hline
\multirow{5}{*}{\bf Swin Transformer  (\cite{swin})}&167 &79.42& 79.41&	52.87& 52.85&	41.94& 41.95\\
&183&79.67& 77.9&	47.3& 47.96&	42.77& 42.67\\
&176 &78.93& 76.23&	49.85& 49.95&	39.56& 39.76\\
&135 &74.49& 74.32&	48.47& 48.57&	38.64& 35.5\\
&119 &71.09& 70.67&	50.34& 51.67&	28.37& 30\\

\hline
\multirow{5}{*}{\bf Mlp-Mixer  (\cite{mlp-mixer})}&167 &68.90&68.54&	39.31& 38.75&	34.67& 33.46\\
&183&69.03& 68.75&	38.71& 38.46&	35.99& 34.94\\
&176 &69.03& 67.45&	40& 39.85&	36.01& 35.84\\
&135 &69.32& 67.29&	40.47& 39.76&	34.62& 34.25\\
&119 &68.93& 67.59&	21.67& 22.45&	25.67& 28\\

\hline
    \end{tabular} 
\end{table}
\subsection{Without weight decay}
\begin{table*}[!htb]
 \caption{Comparision of different baseline models without weight-decay under diverse sampling strategies, including
periodic (Pe), and probabilistic (Pr) samplings. The table includes details of effective back props and total FLOPS (in Tera FLOPS).}
    \label{tab:snodecay}
    \centering
        \tiny

    \begin{tabular}{ll|ccc}
\hline
     {\bf Model}&{\bf Effective epochs}&{\bf CIFAR-10}&{\bf CIFAR-100} & {\bf TINY}  \\ \hline
\multirow{6}{*}{\bf ResNet-50  (\cite{Resnet})}&Baseline (0)&70.34/ 848.76&	32.07/ 848.99&	25.95/ 6759.33\\
&Pr= 5 (20)&69.99&	29.35&	25.2\\
&Pr= 10 (10)&68.32&	31.26&	23.78\\
&Pe= 0.2 (20)&72.05&	31.95&	27.45\\
&Pe= 0.5 (50)&71.29&	31.97&	26.75\\
&Pe= 0.7 (70)&68.96&	31.55&	25.86\\
\hline

\multirow{6}{*}{\bf Swin Transformer  (\cite{swin})}&Baseline (0)&77.92&	45.67&	39.56\\
&Pe= 5 (20)&74.69&45.75&	37.26\\
&Pe= 10 (10)&75.62&	42.67&	39.54\\
&Pr= 0.2 (20)&73.09&	44.69&	32.27\\
&Pr= 0.5 (50)&76.32&	43.27&	31.96\\
&Pr= 0.7 (70)&70.69&	43.96&	31.17\\ \hline

\multirow{6}{*}{\bf MLP-Mixer  (\cite{mlp-mixer})}&Baseline (0)&62.75&	33.26&	30.75\\
&Pe= 5 (20)&60.54&20.76&	28.25\\
&Pe= 10 (10)&61.25&	32.75&	28.42\\
&Pr= 0.2 (20)&62.50&	33.30&	30.8\\
&Pr= 0.5 (50)&62.56&	32.16&	31\\
&Pr= 0.7 (70)&58.46&	28.41&	25.75\\ \hline
\end{tabular}
\end{table*}
From the Table~\ref{tab:snodecay} it was observed that our hypothesis still holds valid with respect to the baseline's performance.

\section{Stochastic Federated Learning (FL) Results}
\label{sec:sFL}
We conducted experiments based on different local epochs.
\begin{table*}
\tiny
    \caption{Evaluating the performance of our Stochastic FL setting upon 20 global rounds about Total Clients (TC), Selected Clients (SC) for each round along with diverse sampling strategies that includes periodic (Pe), probabilistic (Pr) trained over 10 local epochs.}
    \centering
    \begin{tabular}{c|ccl|c|cc|cc|cc|c}
        \hline
        \multirow{2}{*}{\bf Method}&\multirow{2}{*}{\bf TC}& \multirow{2}{*}{\bf SC}&\multirow{2}{*}{\bf Strategy (\% savings)}& \multicolumn{2}{c}{\bf MobileNetV2}& \multicolumn{2}{c}{\bf ShuffleNetV2}& \multicolumn{2}{c}{\bf SqueezeNet}& \multicolumn{2}{c}{\bf ResNet-50} \\ 
        &&&&{\bf CIFAR-10}&{\bf CIFAR-100}&{\bf CIFAR-10}&{\bf CIFAR-100}&{\bf CIFAR-10}&{\bf CIFAR-100}&{\bf CIFAR-10}&{\bf CIFAR-100}\\
        \hline
        \multirow{12}{*}{\tt FedAvg}&\multirow{3}{*}{\bf 5}& \multirow{3}{*}{\bf 1}&Baseline (0)&69.07&	26.16&	64.35&	24.1&	70.72&	28.03&	69.92&	30.01\\
        &&&Pe= 5 (20)&68.76&	25.29&	61.84&	26.04&	63.85&	25.56&	67.37&	25\\
&&&Pe= 10 (10)&67.49&	27.9&	64.48&	26.85&	64.01&	26.22&	69.76&	29.47\\ \cline{2-12}
&\multirow{3}{*}{\bf 5}& \multirow{3}{*}{\bf 2}&Baseline (0)&72.35&	33.16&	67.67&	27.51&	71.66&	31.33&	74&	33.12\\
&&&Pe= 5 (20)&69.16&	30.6&	66.64&	28.25&	66.58&	28.29&	71.46&	31.17\\
&&&Pe= 10 (10)&72.28&	31.27&	68.6&	26.27&	70.61&	27&	75.01&	34.01\\ \cline{2-12}
&\multirow{3}{*}{\bf 10}& \multirow{3}{*}{\bf 2}&Baseline&66.19	&26.54	&62.46	&23.11	&63.85	&23.57	&62.75	&26.2\\
&&&Pe= 5&62.47	&23.39	&58.85	&23.81&	59.58&	19.74&	63.34&	23.63\\
&&&Pe= 10&65.15&	24.51	&60.91	&24.94&	61.55&	20.06&	63.63&	24.59\\ \cline{2-12}
&\multirow{3}{*}{\bf 10}& \multirow{3}{*}{\bf 5}&Baseline (0)&68.2&	27.45	&63.02&	25.18&	63.52&	25&	68.72&	29.55\\
&&&Pe= 5 (20) & 64.8&	25.58&	61.85&	24.05&	59.47&	17.1&	66.27&	27.14\\
&&&Pe= 10 (10)& 68.17&	27.48&	62.9&	25.79&	62.97&	22.43&	68.44&	28.28\\ \hline

\multirow{12}{*}{\tt FedProx($\mu=$ 0.2)}&\multirow{3}{*}{\bf 5}& \multirow{3}{*}{\bf 1}&Baseline (0)& 70.23&	28.33&	66.39&	25.6&	71.69&	29.16&	69.99&	30.99\\
&&&Pe= 5 (20)&	71.86&	27.25&	63.89&	27.03&	65.72	&26.42	&68.15	&26\\
&&&Pe= 10 (10)&	70.53&	28.5&	65.48&	27.55&	66.55&	27.19&	69.81&	29.35\\  \cline{2-12}
&\multirow{3}{*}{\bf 5}& \multirow{3}{*}{\bf 2}&Baseline&	73.45&	35.26&	68.59&	28.42&	71.95&	31.95&	74.25&	34.65\\ 
&&&Pe= 5 (20)&	70.47&	33.25&	67.54&	29.15&	67.25&	29.33&	72.49&	32.99\\
&&&Pe= 10 (10)&	73.15&	34.86&	69.7&	27.99&	69.71&	30.55&	75.12&	35\\  \cline{2-12}
&\multirow{3}{*}{\bf 10}& \multirow{3}{*}{\bf 2}&Baseline (0)&	68.55&	28.66&	63.85&	23.95&	64.26&	25.61&	63.15&	26.51\\
&&&Pe= 5 (20)&	63.75&	25.5&	59.92&	23.82&	60.15&	20.33&	64&	24.15\\
&&&Pe= 10 (10)&	65.25&	26.55&	61.95&	24.99&	63.55&	21.95&	64.25&	24.91\\ \cline{2-12}
&\multirow{3}{*}{\bf 10}& \multirow{3}{*}{\bf 5}&Baseline (0)&	70.5&	29.75&	64.15&	26.07&	64.71&	24.66&	68.99&	30.16\\ 
&&&Pe= 5 (20)&	69.8&	28.66&	63.72&	24.93&	60.51&	19.25&	67.18&	29.84\\
&&&Pe= 10 (10)	&70.35&	29.76&	64.05&	26.79&	63.85&	23&	68.21&	29.65\\ \hline

    \end{tabular}

    \label{tab:sFL_10}
\end{table*}

\begin{table*}
\caption{Evaluating the performance of our Stochastic FL setting upon 30 global rounds about Total Clients (TC), Selected Clients (SC) for each round along with diverse sampling strategies that includes periodic (Pe), probabilistic (Pr) trained over 30 local epochs.}
\tiny
    \centering
    \begin{tabular}{c|ccl|c|cc|cc|cc|c}
        \hline
        \multirow{2}{*}{\bf Method}&\multirow{2}{*}{\bf TC}& \multirow{2}{*}{\bf SC}&\multirow{2}{*}{\bf Strategy (\% savings)}& \multicolumn{2}{c}{\bf MobileNetV2}& \multicolumn{2}{c}{\bf ShuffleNetV2}& \multicolumn{2}{c}{\bf SqueezeNet}& \multicolumn{2}{c}{\bf ResNet-50} \\ 
        &&&&{\bf CIFAR-10}&{\bf CIFAR-100}&{\bf CIFAR-10}&{\bf CIFAR-100}&{\bf CIFAR-10}&{\bf CIFAR-100}&{\bf CIFAR-10}&{\bf CIFAR-100}\\
        \hline
        \multirow{12}{*}{\tt FedAvg}&\multirow{3}{*}{\bf 5}& \multirow{3}{*}{\bf 1}&Baseline (0)&69.5&	28.42&	60.88&	25.41&	73.15&	28.225&	72.775&	33.095\\
        &&&Pe= 5 (20)&67.95&	28.76&	59.26&	25.45&	69.66&	25.82&	68.8&	30.06\\
&&&Pe= 10 (10)&69.28&	27.55&	59.71&	26.47&	70.17&	26.65&	71.5&	32.79\\ \cline{2-12}
&\multirow{3}{*}{\bf 5}& \multirow{3}{*}{\bf 2}&Baseline (0)&74.24&	35.67&	68.46&	30.26&	73.89&	33.3&	75.76&	37.04\\
&&&Pe= 5 (20)&71.92&	30.83&	66.03&	29.66&	70.61&	30.68&	73.31&	34.37\\
&&&Pe= 10 (10)&73.59&	33.13&	66.17&	28.09&	72.76&	29.83&	76.14&	37.49\\ \cline{2-12}
&\multirow{3}{*}{\bf 10}& \multirow{3}{*}{\bf 2}&Baseline (0)&67.05&	26.87&	60.22&	23.33&	70.03&	25.49&	66.03&	29.4\\
&&&Pe= 5 (20)&61.88&	23.81&	57.94&	22.66&	66.02&	22.9&	64.79&	26.53\\
&&&Pe= 10 (10)&67.16&	25.66&	60.6&	24.48&	68.53&	22.52&	65.26&	28.01\\ \cline{2-12}
&\multirow{3}{*}{\bf 10}& \multirow{3}{*}{\bf 5}&Baseline (0)&69.8&	27.55&	63.04&	27.06&	70.72&	28.18&	69.13&	32.76\\
&&&Pe= 5 (20) & 66.67&	26.19&	62.1&	26.06&	67.75&	23.16&	67.55&	29.49\\
&&&Pe= 10 (10)& 69.36&	28.4&	62.94&	27.63&	70.62&	26.29&	69.23&	30.68\\ \hline

\multirow{12}{*}{ \tt FedProx($\mu=$ 0.2)}&\multirow{3}{*}{\bf 5}& \multirow{3}{*}{\bf 1}&Baseline (0)& 70.82&	30.42&	62.29&	26.37&	74&	28.92&	72.95&	34.05\\
&&&Pe= 5 (20)&	70.05&	28.88&	60.89&	26.32&	70.98&	26.66&	69.11&	31.05\\
&&&Pe= 10 (10)&	71.31&	29.2&	61.16&	27.28&	72.5&	27.59&	71.83&	33.2\\  \cline{2-12}
&\multirow{3}{*}{\bf 5}& \multirow{3}{*}{\bf 2}&Baseline (0)&	74.8&	37.25&	69.33&	31.21&	75&	34.09&	76.45&	37.85\\ 
&&&Pe= 5 (20)&	72.73&	32.7&	67.89&	31.12&	71.45&	31.78&	74.2&	35.52\\
&&&Pe= 10 (10)&	74.1&	33.95&	68.57&	30.25&	72.85&	31.81&	76.43&	38\\  \cline{2-12}
&\multirow{3}{*}{\bf 10}& \multirow{3}{*}{\bf 2}&Baseline (0)&	68.73&	28.91&	62.05&	24.04&	70.58&	26.63&	66.85&	29.99\\
&&&Pe= 5 (20)&	64.93&	26.47&	59.97&	22.93&	67.4&	23.57&	66.92&	27.75\\
&&&Pe= 10 (10)&	67.15&	27.31&	61.6&	25&	70.61&	23.69&	66.64&	28.78\\ \cline{2-12}
&\multirow{3}{*}{\bf 10}& \multirow{3}{*}{\bf 5}&Baseline (0)&	71.27&	29.83&	63.85&	27.53&	71.35&	28.44&	69.97&	34.17\\ 
&&&Pe= 5 (20)&	69.47&	28.4&	63.36&	26.19&	68.71&	24.95&	68.16&	31.29\\
&&&Pe= 10 (10)	&70.8&	30.35&	63.9&	28.44&	71.42&	26.08&	69.6&	33.15\\ \hline

    \end{tabular}
    
    \label{tab:sFL_30}
\end{table*}

\newpage
\section{Domain Adaptation (DA) and Domain Generalization (DG)}
\label{sec:sdadg}
\textbf{Domain Adaptation (DA)} is a machine learning technique that aims to find a common representation between the \textbf{"source" and "target" domains}. In contrast, \textbf{Domain Generalization} seeks to find a common representation across \textbf{multiple source domains and an unseen target domain}.

\par {\bf Datasets:} Office-31  (\cite{office31}) is a standard domain adaptation dataset comprising three diverse domains: Amazon (from the Amazon website), Webcam (captured by web cameras), and DSLR (captured by digital SLR cameras), with a total of 4,652 images across 31 unbalanced classes. VisDA  (\cite{visda}) is a simulation-to-real dataset featuring two extremely distinct domains: Synthetic renderings of 3D models and Real images collected from photo-realistic or real-image datasets. With 280K images in 12 classes, the scale of VisDA-2017 presents significant challenges to DA. For DG, we experimented with Office-Home  (\cite{officehome}), which contains domains such as Art, Clipart, Product, and Real-World, and PACS  (\cite{pacs}), which includes domains like Photo, Art, Cartoon, and Sketch.

\par {\bf Methods:} We conducted experiments using various standard domain adaptation (DA) measurements, including Marginal Disparity Discrepancy (MDD)  (\cite{MDD}), which introduces a marginal loss to provide generalization bounds for multiclass DA. We also employed Minimum Class Confusion (MCC)  (\cite{MCC}), which minimizes between-class confusion while maximizing within-class confusion to enhance transfer gains. For domain generalization (DG), we experimented with VREx  (\cite{VREx}), aimed at reducing a model’s sensitivity to extreme distributional shifts by optimizing over extrapolated domains and penalizing training risk variance. Additionally, we utilized GroupDRO  (\cite{GroupDRO}), a method that applies regularized group distributionally robust optimization to improve worst-case performance of overparameterized neural networks, thus enhancing generalization on atypical data groups.

\begin{table*}
\caption{Performance across individual classes and overall average (Avg (\%)) on distinct domains to Synthetic domain DA methods over VisDA  (\cite{visda}) dataset.}
    \tiny
    \centering
    \begin{tabular}{cl|cccccccccccc|c}
    \hline
         {\bf Method}&{\bf Strategy (\% savings)}&{\bf Aeroplane}&	{\bf Bicycle}&	{\bf Bus}&	{\bf Car}&	{\bf Horse}&	{\bf Knife}&	{\bf Motorcycle}&	{\bf Person}&	{\bf Plant}&	{\bf Skateboard}&	{\bf Train}&	{\bf Truck}&	{\bf Avg}  \\ \hline
         \multirow{6}{*}{\bf MDD  (\cite{MDD})}& Baseline (0)&95.09&	60.8&	82.6&	70.37&	89.57&	47.46&	91.66&	78.94&	89.33&	74.09&	84.65&	42.57&	75.6\\
&Pe= 5 (20)&94.29&	70.38&	84.15&	67.45&	90.81&	12.72&	91.51&	81.19&	88.74&	88.07&	83.73&	39.31&	74.37\\
&Pe= 10 (10)&94.84&	68.74&	81.04&	66.23&	92.17&	59.22&	91.09&	81.72&	86.61&	70.97&	84.13&	41.74&	76.54\\
&Pr= 0.2 (20)&94.24&	61.32&	78.59&	69.52&	89.1&	17.68&	91.49&	75.17&	93.51&	79.39&	81.3&	38.44&	72.48\\
&Pr= 0.5 (50)&93.99&	70.04&	75.86&	64.08&	86.69&	14.26&	90.16&	75.82&	84.17&	81.07&	83.47&	44.43&	72\\
&Pr= 0.7 (70)&92.37&	61.17&	80.66&	63.33&	86.78&	19.95&	88.26&	75.55&	78.76&	84.56&	80.09&	43.88&	71.28\\
\hline
\multirow{6}{*}{\bf MCC  (\cite{MCC})}& Baseline (0)&95&	82.6&	72.06&	65.53&	90.47&	15.61&	82.57&	79.14&	88.74&	84.83&	84.3&	54.05&	74.71\\
&Pe= 5 (20)&95.11&	83.68&	75.33&	61.57&	91.17&	15.51&	83.28&	78.25&	87.18&	85.88&	85.62&	55.87&	74.88\\
&Pe= 10 (10)&95.14&	83.56&	75.33&	61.65&	90.83&	14.89&	82.26&	77.62&	86.96&	86.89&	86.16&	55.42&	74.72\\
&Pr= 0.2 (20)&95.36&	83.42&	76.41&	63.2&	90.76&	15.9&	82.62&	78.02&	85.95&	86.23&	84.51&	54.21&	74.71\\
&Pr= 0.5 (50)&94.84&	82.84&	75.35&	63.53&	90.85&	14.45&	84.36&	77.15&	86.87&	84.52&	85.38&	52.64&	74.4\\
&Pr= 0.7 (70)&94.51&	81.41&	76.22&	63.89&	90.49&	13.01&	86.11&	76.5&	87.49&	83.77&	84.25&	51.55&	74.1\\ \hline
         
    \end{tabular}
    
    \label{tab:DA_VisDA}
\end{table*}

\begin{table}[!ht]
\tiny
\caption{Performance across various Domain Generalization (DG) methods on the PACS  (\cite{pacs}) dataset.}
    \label{tab:DG-pacs}
    \centering
    \begin{tabular}{cl|cccc|c}
    \hline
    {\bf Method}  &{\bf Strategy (\% savings)} &{\bf A}&{\bf C}&{\bf P}&{\bf S}&{\bf Avg}  \\ \hline
      \multirow{6}{*}{\bf VREx  (\cite{VREx})}&Baseline (0)&	87.2&	82.3&	97.4&	81	&87.0 \\
&Pe = 5 (20)&	87&	82&	97&	80.8&	86.7 \\
&Pe = 10 (10)&	87.5&	82.5&	97.5&	81.2&	87.2 \\
&Pr = 0.2 (20)&	87.2&	81.9&	97.8&	81.1&	87.0 \\
&Pr = 0.5 (50)&	86.5&	82&	97.3&	80	&86.5 \\
&Pr = 0.7 (70)&	87&	82&	97&	79.5&	86.4 \\ \hline

\multirow{6}{*}{\bf GroupDRO  (\cite{GroupDRO})}&Baseline (0)&	88.9&	81.7&	97.8&	80.8&	87.3 \\
&Pe = 5 (20)&	88.5&	81&	97.5&	80.5&	86.9 \\
&Pe = 10 (10)&	89&	81.7&	98&	81&	87.4 \\
&Pr = 0.2 (20)&	89&	82&	97.2&	80.9&	87.3 \\
&Pr = 0.5 (50)&	88.2&	81.2&	97.9&	80.5&	87\\
&Pr = 0.7 (70)&	87.9&	80.9&	96&	79&	86 \\
        \hline       
    \end{tabular}
\end{table}
\end{document}